\documentclass[preprint, 3p,review, table]{elsarticle}
\biboptions{numbers}
\usepackage{setspace}
\usepackage{subfiles}
\usepackage{etoolbox}
\usepackage{amssymb}
\usepackage{amsmath,amssymb}
\usepackage{dutchcal}
\patchcmd{\maketitle}% <cmd>
  {\finalMaketitle\tableofcontents\vspace{\baselineskip}}
  {}{}% <success><failure>

\usepackage{framed} % Framing content 
\usepackage{multicol} % Multiple columns environment
\usepackage{multirow}
\usepackage{nomencl} % Nomenclature package
\usepackage{comment}
\usepackage{placeins}
\usepackage{colortbl}
\usepackage{soul}  
\sethlcolor{white}
\usepackage{pdflscape}
\usepackage{subcaption} 
\usepackage[ruled,vlined]{algorithm2e}
\usepackage[table,xcdraw]{xcolor}
\usepackage{babel}
\usepackage[numbers]{natbib}
\makenomenclature
\usepackage{etoolbox}
\renewcommand\nomgroup[1]{%
  \item[\Large\bfseries
  \ifstrequal{#1}{N}{Nomenclature}{%
  \ifstrequal{#1}{P}{Parameters}{}}%
]\vspace{10pt}} % This is to add vertical space between the groups.
\renewcommand*\nompreamble{\begin{multicols}{2}} 
\renewcommand*\nompostamble{\end{multicols}}
\usepackage{longtable}
\usepackage{pdflscape} % For landscape environment
\usepackage{booktabs} % For better table rules
\usepackage{multirow} % For multi-row cells
\usepackage{float}
\usepackage{enumerate}
\usepackage{enumitem}
\usepackage{graphicx}
\usepackage{tabularx}
\usepackage{multirow}
\usepackage{geometry}
\usepackage{amsmath}
\usepackage{amssymb}
\usepackage{xcolor}
\usepackage{amsfonts,amsthm,bm} % Math packages
\usepackage{color,soul}
\usepackage{colortbl,booktabs}
\usepackage{enumerate}
\usepackage{amsmath}
\usepackage[version=4]{mhchem}  % Load the mhchem package
\usepackage{tikz}
\usepackage{gensymb}
\usepackage{adjustbox} % for shifting the tikzpicture
\usetikzlibrary{mindmap}
\floatstyle{plaintop} % This line and the next are to move the table caption to the top of the table.
\usepackage[table,xcdraw]{xcolor}  % Make sure you include this package

\restylefloat{table}
\usepackage{rotating}
\usepackage{smartdiagram}
\usesmartdiagramlibrary{additions}
\usepackage{caption}
\usepackage{subcaption}
\biboptions{numbers,compress}
\usepackage{hyperref}
\usepackage{url}
\hypersetup{
    colorlinks=true, %set true if you want colored links
    linktoc=all,     %set to all if you want both sections and subsections linked
    linkcolor=blue,  %choose some color if you want links to stand out
    citecolor=blue,  %choose some color if you want references to stand out
}
\usepackage{tikz}
\usetikzlibrary{mindmap}
\usetikzlibrary{trees}
\usetikzlibrary{shadings}
\tikzstyle{every node}=[draw=black,thin,anchor=west, minimum height=2.5em]

\renewcommand\paragraph{\@startsection{paragraph}{4}{\z@}%
            {-2.5ex\@plus -1ex \@minus -.25ex}%
            {1.25ex \@plus .25ex}%
            {\normalfont\normalsize\bfseries}}
\long\def\symbolfootnote[#1]#2{\begingroup%
\def\thefootnote{\fnsymbol{footnote}}\footnote[#1]{#2}\endgroup}
\usepackage{sectsty} % to change section font size
\sectionfont{\fontsize{12}{14}\selectfont}

\definecolor{main}{RGB}{11,79,108}
\definecolor{c1}{RGB}{252,213,129}
\definecolor{c2}{RGB}{255,155,113}
\definecolor{c3}{RGB}{234,218,162}
\definecolor{c4}{RGB}{67,129,193}
\definecolor{c5}{RGB}{134,222,183}
\definecolor{c6}{RGB}{184,140,167}
\definecolor{c7}{RGB}{102,137,161}
\definecolor{c8}{RGB}{132,230,248}
\definecolor{c9}{RGB}{200,159,156}
\definecolor{c10}{RGB}{138,162,158}
\definecolor{c11}{RGB}{195,190,247}
\definecolor{c12}{RGB}{16,115,158}
\definecolor{c13}{RGB}{0,0,0}
\graphicspath{{./}}

\tikzset{
  treenode/.style = {shape=rectangle,
                     draw, align=center,
                     top color=white, bottom color=blue!20},
  root/.style     = {treenode, font=\Large, bottom color=red!30},
  env/.style      = {treenode, font=\normalsize},
  dummy/.style    = {circle,draw}
}

\nomenclature[N]{GHG}{Green House Gas}
\nomenclature[N]{RL}{Reinforcement Learning}
\nomenclature[N]{MCE}{Markov Chain Evolution}
\nomenclature[N]{MTB}{Micro-Trip-Based}
\nomenclature[N]{MCB}{Marcov-Chain-Based}
\nomenclature[N]{TPM}{Transition Probability Matrix}
\nomenclature[N]{PHEV}{Plug-in Hybrid Electric Vehicle}
\nomenclature[N]{OBD}{On-Board Diagnostics}
\nomenclature[N]{GPS}{Global Positioning System}
\nomenclature[N]{ECU}{Electronic Control Unit}
\nomenclature[N]{CAN}{Controller Area Network}
\nomenclature[N]{MAE}{Mean Absolute Error}
\nomenclature[N]{MSE}{Mean Squared Error}
\nomenclature[N]{RMSE}{Rooted Mean Squared Error}
\nomenclature[N]{SAGSTM}{Speed Acceleration Grade State Transition Matrix}
\nomenclature[N]{MDP}{Markov Decision Process}
\nomenclature[N]{TD}{Temporal Difference}
\nomenclature[N]{VSP}{Vehicle Specific Power}
\nomenclature[N]{IMU}{Inertial Measurement Unit}
\nomenclature[N]{PIESMC}{Physics Informed Expected SARSA-Monte Carlo}
\nomenclature[N]{RDCC}{Representative Driving Cycle Construction}
\nomenclature[N]{MOVES}{MOtor Vehicle Emission Simulator}

\nomenclature[P]{$\alpha_{ES}$}{Learning Rate for Expected SARSA Method}
\nomenclature[P]{$\alpha_{MC}$}{Learning Rate for Monte Carlo Method}
\nomenclature[P]{$\gamma_{ES}$}{Discount Factor for Expected SARSA Method}
\nomenclature[P]{$\gamma_{MC}$}{Discount Factor for Monte Carlo Method}
\nomenclature[P]{$\delta$}{Temporal Difference Error}
\nomenclature[P]{$\tau$}{Temperature parameter}
\nomenclature[P]{$\beta$}{Scaling factor}
\nomenclature[P]{$\lambda_{ext}$}{Extrinsic reward weight}
\nomenclature[P]{$\lambda_{int}$}{Intrinsic rewards weight}
\nomenclature[P]{$\varsigma$}{scaling factor}
\nomenclature[P]{$w_{ES}$}{State-action value's weight for Expected SARSA Method}
\nomenclature[P]{$w_{MC}$}{State-action value's weight for Monte Carlo Method}
\nomenclature[P]{$\epsilon$}{Epsilon value}

\journal{Transportation Research Part D: Transport and Environment}
\begin{document}

\begin{frontmatter}

\title{A Generative Physics-Informed Reinforcement Learning-Based Approach for Construction of Representative Drive Cycle.}
\author[1]{Amirreza Yasami}
\ead{ayasami@ualberta.ca}
\author[1]{Mohammadali Tofigh}
\author[1]{Mahdi Shahbakhti}
\author[1]{Charles Robert Koch}

\affiliation[1]{organization={Mechanical Engineering Department, University of Alberta},
            addressline={116 St and 85 Ave},
            city={Edmonton},
            postcode={T6G 1H9},
            state={Alberta},
            country={Canada}}

\begin{abstract}

Accurate driving cycle construction is crucial for vehicle design, fuel economy analysis, and environmental impact assessments. A generative Physics-Informed Expected State Action Reward State Action (SARSA)-Monte Carlo (PIESMC) approach that constructs representative driving cycles by capturing transient dynamics, acceleration, deceleration, idling, and road grade transitions while ensuring model fidelity is introduced. Leveraging a physics-informed reinforcement learning framework with Monte Carlo sampling, PIESMC delivers efficient cycle construction with reduced computational cost. Experimental evaluations on two real-world datasets demonstrate that PIESMC replicates key kinematic and energy metrics, achieving up to a 57.3\% reduction in cumulative kinematic fragment errors compared to the Micro-trip-based (MTB) method and a 10.5\% reduction relative to the Markov-chain-based (MCB) method. Moreover, it is nearly an order of magnitude faster than conventional techniques. Analyses of vehicle-specific power distributions and wavelet-transformed frequency content further confirm its ability to reproduce experimental central tendencies and variability.

\end{abstract}

\end{frontmatter}

%%%%%%%%%%%%%%%%%%%%%%%%%%%%%%%%%%%%%%%%%%%%%%%%%% nomenclature
% \begin{table*}[ht!]
% \begin{framed}
% \begin{footnotesize}
% \printnomenclature
% \end{footnotesize}
% \end{framed}
% \end{table*}
% %%%%%%%%%%%%%%%%%%%%%%%%%%%%%%%%%%%%%%%%%%%%%%%%%%

% \tableofcontents

\section{Introduction} \label{sec:intro}

\subsection{Background and related works}

Climate change is one of the most significant challenges that humanity is facing, whose detrimental impacts extend beyond the environment and natural resources, leading to health and economic challenges~\cite{canadaGHE}. Fossil fuels, such as coal, oil, and gas, are the primary contributors to climate change, responsible for over 75\% of global greenhouse gas (GHG) emissions and nearly 90\% of all carbon dioxide emissions~\cite{UNClimateChange}.
The global transportation sector accounts for 7943 Mt \(CO_2\) eq of global GHG emissions, representing 20\% of the total GHG emissions in 2022. This reflects a 22\% increase in global transportation GHG emissions since 2005~\cite{EDGAR2024}. In Canada, the transportation sector contributes 28\% of the nation’s total GHG emissions, and it has been considered the second largest contributor to the total GHG emissions. Moreover, trends indicate an increase of 5 Mt \(CO_2\)  eq (3\%) in transportation emissions from 2005 to 2022~\cite{ECCC2024}. 

To mitigate these emissions, many governments worldwide have implemented stringent policies and actions, such as establishing emission standards and test procedures for on-road vehicles and engines aimed at reducing fuel consumption and emissions in vehicles \cite{EPAVehicleRegulations, CanadaRegulations2024}. These regulations and government initiatives to reduce emissions have led to research and development in areas related to the development of advanced automobile technologies and new approaches for emission monitoring and estimation. These research efforts enable the assessment of various scenarios involving new and emerging vehicle technologies, as well as other factors influencing emission inventories.

Reliable representative driving cycles serve as a foundation for accurately analyzing and predicting vehicle emissions and energy consumption under various conditions~\cite{pan2021fuzzy, huzayyin2021representative}. The standard certification driving tests do not properly characterize the real driving conditions \cite{fontaras2017fuel}; therefore, there will be a difference in the energy consumption and emission production values reported by the certification tests compared to those in real driving conditions. The factors that have an impact on this difference have been analyzed in \cite{fontaras2017fuel}. Furthermore, having a local driving cycle is crucial for accurate energy consumption and emission estimation as it is specific to each area or region \cite{alam2017modeling}. It has been demonstrated that the driving patterns are different from one city to another \cite{kamble2009development, yuhui2019development}. Therefore, studies have been conducted worldwide to generate local driving cycles \cite{ashtari2014using, giakoumis2017driving}. 

The existing methods in the literature for the generation of representative driving cycles are mostly based on stochastic approaches, including micro-trip-based (MTB) techniques, also known as segmentation methods, cluster-based approaches, and Markov chain-based (MCB) methods \cite{quirama2020driving, giraldo2021effect, forster2019data, sundarkumar2021time, yuhui2019development}.

MTB, a primary approach for drive cycle development, involves segmenting the real trip data from the tested vehicles~\cite{wang2008characterization, lipar2016development}. Segments can be defined in various ways, such as micro-trips, portions between stops, or based on operating modes, acceleration, deceleration, and cruising. Then, a representative driving cycle will be constructed by random sampling. The major advantage of the MTB method is its simplicity. However, the MTB has limitations in scenarios with rare stops, such as highway driving~\cite{cui2021optimization}, and in determining a representative driving cycle because of its uncertainty~\cite{chen2019optimization}. Furthermore, the MTB method constructs representative driving cycles by combining individual micro-trips, which often results in discontinuous acceleration and grade profiles in the generated driving cycle~\cite{zhang2019using}. Lastly, the constructed representative driving cycle often fails to accurately capture transient sections~\cite{lin2002exploratory}.

A second main approach for constructing a representative driving cycle involves cluster-based methods, where micro-trips are initially grouped into clusters based on their kinematic characteristics. Then, within each cluster, the micro-trip closest to the cluster center is selected~\cite{yang2018markov}. The selected micro-trips are then processed using the same methodology as the MTB approach to construct the representative driving cycle. The driving cycle is then constructed by proportionally selecting micro-trips from each cluster. However, finding the correct number of clusters is challenging~\cite{peng2020driving}, with k-means clustering being implemented to cluster the micro-trips~\cite{sundarkumar2021time}. The key advantage of this method lies in its semi-random nature, achieved through the clustering process \cite{qiu2022clustering}.

A third approach, MCB (or 2D MCB), constructs driving cycles based on a transition probability matrix (TPM) that represents the probabilities of transitions between states~\cite{vskugor2016delivery, nyberg2014generation}. These states can be defined on a second-by-second basis or categorized into different operational modes, such as acceleration, deceleration, cruising, etc. The TPM is derived from the transitions observed in experimental data. Using Monte Carlo sampling, the MCB method connects each state to the next, thereby generating driving cycles. As a result, the constructed driving cycle preserves the complete statistical information of the original dataset~\cite{cui2022novel}. However, accurately defining the state space requires time characterization of speed and acceleration, resulting in an extensive state space. This leads to a large number of Markov chains due to the combinatorial nature of the Monte Carlo random sampling. So, constructing a representative driving cycle is complex and time-consuming. Although incorporating road grade into the state space further increases the complexity and difficulty, \cite{jia2021constructing} developed representative driving cycles using a 2D Markov Chain method, incorporating road grade via average speed-based segment matching. This approach is limited by segment-matching errors and its inability to fully capture road grade transitions present in experimental data. 

Each of the above methodologies has limitations. MTB methods, while commonly used, often fail to accurately capture the transient driving characteristics that significantly impact emissions and energy consumption and lead to discontinuous acceleration and grade profiles~\cite{lin2002exploratory,zhang2019using}. The second approach, clustering-based methods, similarly suffers the same limitation as the MTB approaches, except that the micro-trips are clustered first and then selected proportionally. The MCB method addresses the issues related to the MTB approaches but is computationally expensive and becomes even more so when incorporating road grade~\cite{jia2021constructing}.

Since road grade significantly impacts vehicle fuel consumption and emissions~\cite{zeng2015parallel}, neglecting it leads to inaccuracies in estimating fuel consumption and emissions along driving cycles~\cite{rosero2021effects, he2022impacts, jia2021constructing}. It has been shown that road grade accounts for 1 to 9\% of fuel consumption of commercial vehicles on average and up to 40\% on select real-world driving cycles~\cite{lopp2015evaluating}. Only a limited number of studies have considered this factor in developing driving cycles. For instance, \cite{zhang2021using} has transformed the multi-parameter driving cycle, including velocity, acceleration, and grade, into driving cycles with zero grade. The transformation maintains the engine power, and the Markov Chain Evolution (MCE) method developed by \cite{zhang2018high} has been used for representative driving cycle construction. This method has some drawbacks, such as the simplified grade representation, complexity in practical implementation, and low-speed data is neglected. 

This paper introduces a new strategy based on reinforcement learning to generate a local representative driving cycle that takes the effect of road grade into account. To validate the proposed methodology, a substantial dataset of on-road driving data has been gathered, covering a range of road grades. The primary contributions of this study are as follows:

\begin{itemize}
    \item Extensive on-road real driving with 146 datasets were collected from a PHEV Ford Escape along a specified route, encompassing two different driving behaviors. The data from this study is shared in a publicly accessible repository to promote further research.
    \item A reinforcement learning (RL)-based methodology is developed to generate 3D representative driving cycles, including speed, acceleration, and road grade, providing a more comprehensive representation of driving conditions.
    \item The RL-based method is benchmarked against established methodologies for representative driving cycle development and demonstrates improved performance in capturing transient behaviors with enhanced computational efficiency on real-world data. 
\end{itemize}

The rest of this paper is organized into sections: 
The datasets used in this study are described in Section~\ref{data_desc}. The methodology employed is described in Section~\ref{method}. The results are described in  Section~\ref{results}, and the key findings and future directions are summarized in Section~\ref{conlusion}.

\section{Dataset Description}\label{data_desc}
A detailed description of the data collected for this study is provided in this section. The dataset was collected using a 2021 Ford Escape Plug-in Hybrid Electric Vehicle (PHEV) from the University of Alberta’s fleet. The data acquisition was performed via an onboard diagnostics (OBD-II) device connected to the vehicle’s OBD port, enabling the collection of Electronic Control Unit (ECU) parameters and GPS data. A total of 146 tests were conducted along the predetermined routes shown in Figure~\ref{fig:driving_cycle}. Out of 146 datasets, 105 of them were conducted under conditions of sunny weather with above zero temperatures (Temp$>0^\circ$C) (classified as Driving Cycle 2, DC2), and the remaining tests were collected under snowy conditions with above zero temperatures (Temp$>0^\circ$C) (defined as Driving Cycle 1, DC1).

\subsection{Test Routes and Data Preprocessing}  
To capture both urban driving behavior and validate grade extraction, two distinct test routes were selected: an urban loop (Route A) and a rural grade loop (Route B).  

Route A covers approximately 20.5 km around the University of Alberta’s North and South campuses, encompassing 31 signalized intersections, 13 stop‐sign intersections, and six pedestrian crossings (Figure~\ref{fig:driving_cycle}). Speeds range from 0 to 28 m/s (100 km/h), ensuring representation of residential streets, arterial roads, and a short highway segment. We deliberately chose this loop to capture stop‐and‐go phases (70\% of total distance), frequent accelerations/decelerations, and mixed traffic densities. All 30 Hz on‐board measurements (PEMS and CAN) from these runs were combined to form the “urban dataset,” from which we derived speed–acceleration–grade clusters for cycle synthesis.  

Route B is a 15 km rural loop selected for its pronounced elevation changes: three uphill segments (3 – 5\% grade), two downhill segments (–3\% to –5 \% grade), and several flat sections (0 – 1\% grade) (Figure~\ref{fig:grade_validation}). By traversing lightly trafficked roads with reliable GPS reception, we minimized speed fluctuations unrelated to grade. Altitude readings from the GPS logger were compared to a high‐resolution Digital Elevation Model (DEM) to compute a reference grade profile. After applying a fourth‐order Butterworth low‐pass filter (cut‐off = 0.1 Hz), this “trusted” grade signal was used to calibrate smoothing parameters applied to Route A’s GPS data. Any systematic bias detected on Route B (e.g., 0.2\% underestimation of uphill grade) was corrected in the urban grade profiles prior to cycle synthesis.  

\begin{figure}[htbp]
  \centering
  \setlength{\abovecaptionskip}{0pt}
  \begin{tabular}{@{}cc@{}}
    \subcaptionbox{\label{fig:driving_cycle}}[0.30\textwidth]{%
      \includegraphics[width=0.32\textwidth]{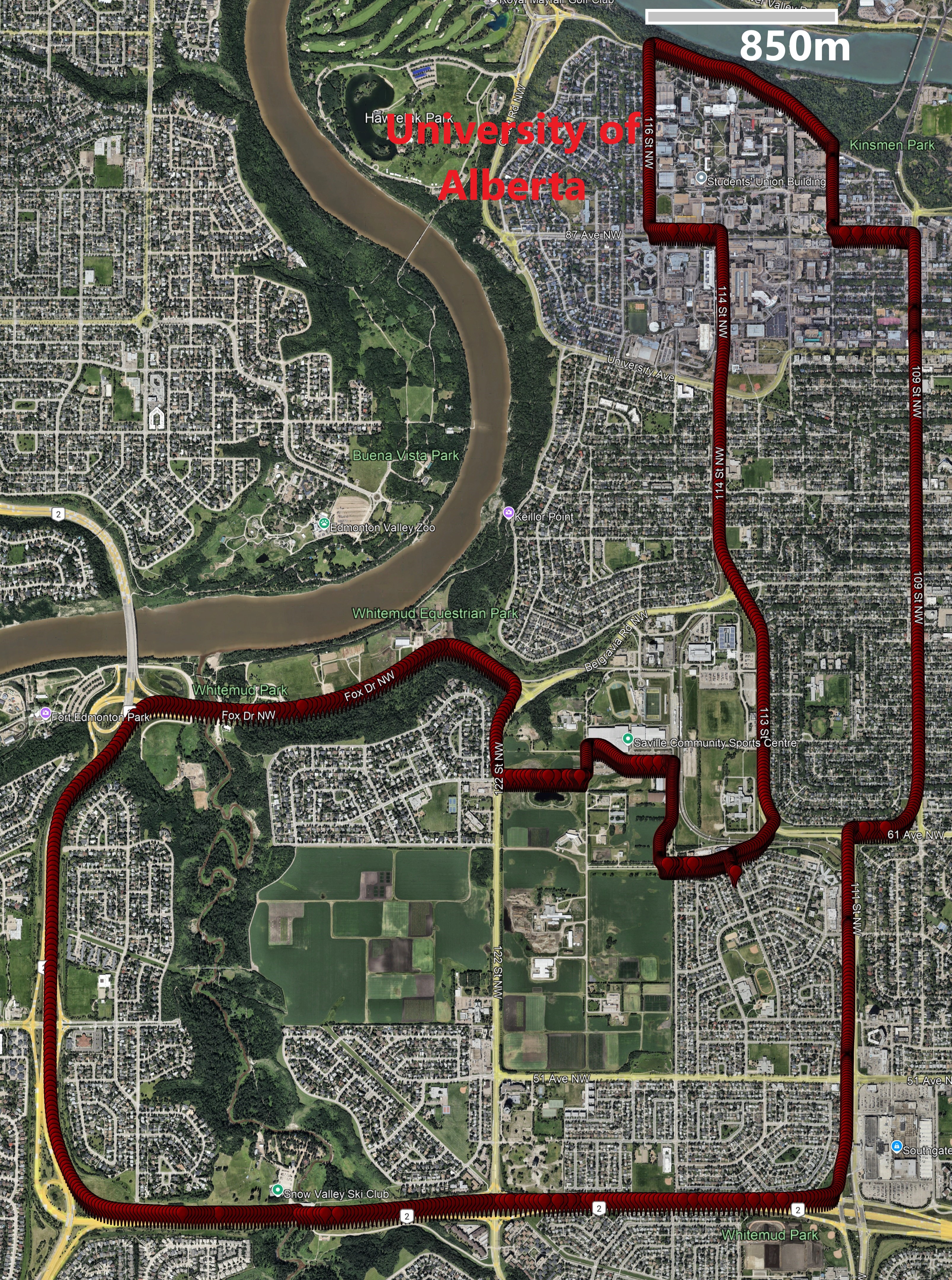}%
    } &
    \subcaptionbox{\label{fig:grade_validation}}[0.30\textwidth]{%
      \includegraphics[width=0.32\textwidth]{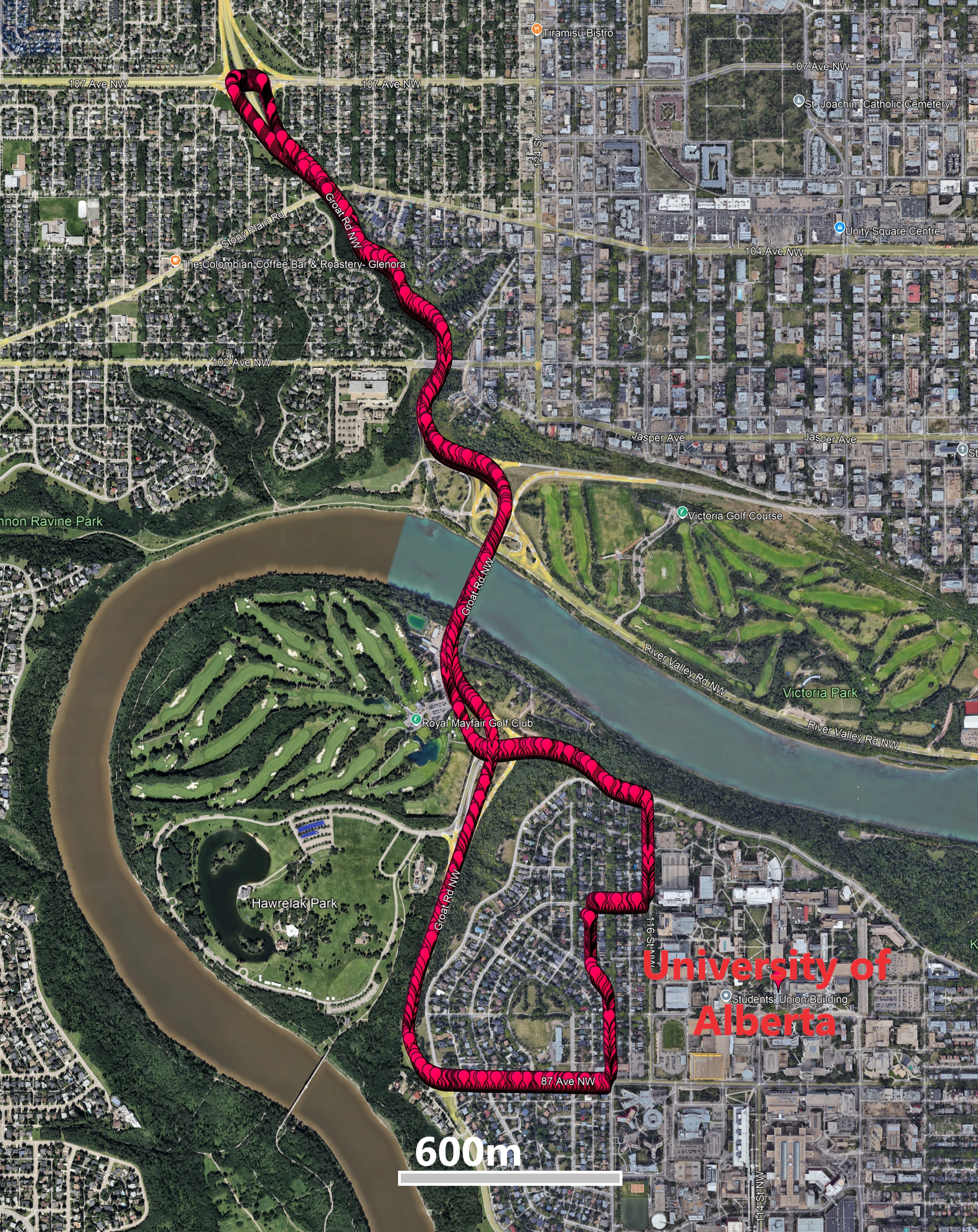}%
    }\\ 
  \end{tabular}
  \caption{Geographical test routes: (a) urban driving cycle test route (Route A) and (b) grade validation route (Route B).}
  \label{fig:urban_route}
\end{figure}

\paragraph{Data Preprocessing Workflow}  
All raw measurements (30 Hz) underwent the following steps:  
\begin{enumerate}
  \item \textbf{Interpolation and Outlier Correction:}  
    \begin{itemize}[leftmargin=1em]
      \item Missing or erroneous speed values (outside [0 m/s, 28 m/s] on Route A and [0 m/s, 22 m/s] on Route B) were linearly interpolated over a 2 s window. Negative or implausible values were clipped to the nearest feasible bound.  
      \item GPS points deviating more than 15 m from the mapped Route A alignment were discarded; on Route B, GPS dropouts longer than 5 s were replaced using DEM‐derived altitudes.  
    \end{itemize}
  \item \textbf{Acceleration Computation:}  
    Longitudinal acceleration was computed via the central‐difference method applied to corrected speed data.
  \item \textbf{Grade Extraction and Smoothing:}  
    \begin{itemize}[leftmargin=1em]
      \item \textit{Route B:} Raw GPS altitude was compared against the DEM (10 m resolution) to quantify measurement error. The resulting difference informed the design of a fourth‐order Butterworth low‐pass filter (cut‐off = 0.1 Hz), yielding a “reference grade” time series.  
      \item \textit{Route A:} The same filter parameters were then applied to Route A’s GPS altitude data, producing a smoothed grade profile used in the speed–acceleration–grade state‐transition matrix (SASTM). Any per‐bin grade offset detected on Route B was imposed on Route A’s grade data before cycle synthesis.  
    \end{itemize}
  \item \textbf{Resampling:}  
    To reduce computational load without altering kinematic statistics (validated by comparing 30 Hz vs. 1 Hz statistics with < 1\% error), all variables were downsampled to 1 Hz.
\end{enumerate}

By explicitly selecting and validating these two routes—Route A for representative urban dynamics and Route B for grade fidelity—and by applying a uniform preprocessing pipeline (interpolation, outlier correction, acceleration computation, filter calibration, and resampling), the final dataset accurately reflects both traffic‐induced behavior and terrain‐induced grade variations.

\subsection{Test Routes and Data Preprocessing} 
The urban route, illustrated in Figure~\ref{fig:driving_cycle}, primarily traverses residential and urban areas. It features 31 intersections with traffic lights, 13 intersections with stop signs, and six pedestrian crossing signals. The route passes through both the University of Alberta's North and South Campuses. The speed along the route ranges from 0 to 28m/s (approximately 100km/h), indicating that the route encompasses a diverse range of road types, including urban streets and highways. The total distance of the route is approximately 20.5km, typically taking around 36 minutes to complete. Urban driving conditions account for 70\% of the total distance traveled.

\begin{figure}[htbp]
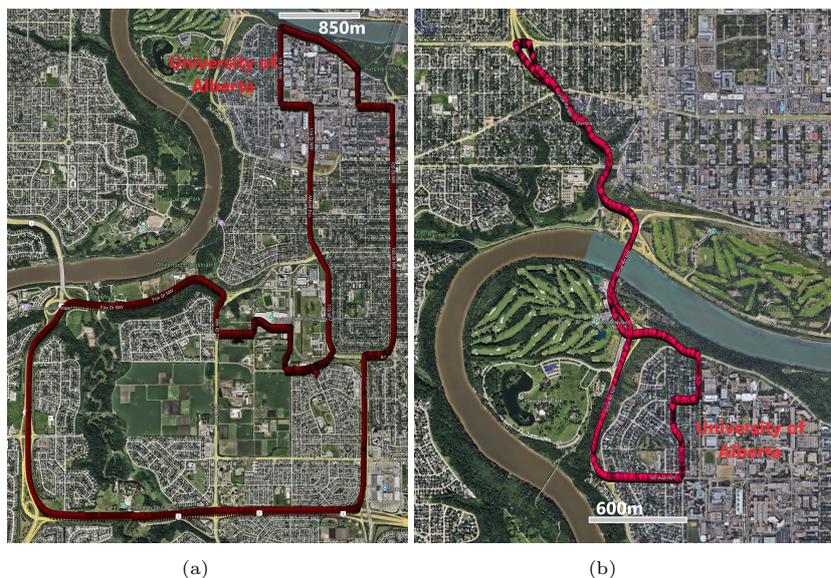

  \centering
  \setlength{\abovecaptionskip}{0pt} % Adjust space between caption and the figure
  \begin{tabular}{@{}cc@{}}
    \subcaptionbox{\label{fig:driving_cycle}}[0.30\textwidth]{\includegraphics[width=0.32\textwidth]{Figures/Test_Route_Ualberta.jpg}} &
    \subcaptionbox{\label{fig:grade_validation}}[0.30\textwidth]{\includegraphics[width=0.342\textwidth]{Figures/Road_Grade_Validataion_Route.jpg}}\\ 
  \end{tabular}
  \caption{The geographical test routes. a) the urban driving cycle test route and b) the grade validation test route.}
  \label{fig:urban_route}
\end{figure}

A series of preprocessing steps were implemented to ensure a clean and reliable dataset. Missing data were first interpolated, and measurements such as speed values that exceeded vehicle thresholds or were negative were corrected using realistic estimates, while GPS points that significantly deviated from the prescribed route were discarded. Acceleration was computed from the speed data via the central difference method. Then, the grade profile was derived from GPS data (latitude, longitude, and altitude) and smoothed with a low-pass filter to mitigate noise. Finally, to optimize computational efficiency without compromising analytical rigor, the original 30Hz dataset was resampled to 1Hz. This adjustment does not affect the generalization of the methods or the validity of the results.

\subsection{Road Grade Calculation}
\begin{figure}[htbp]
  \centering
  \includegraphics[width=0.80\textwidth]{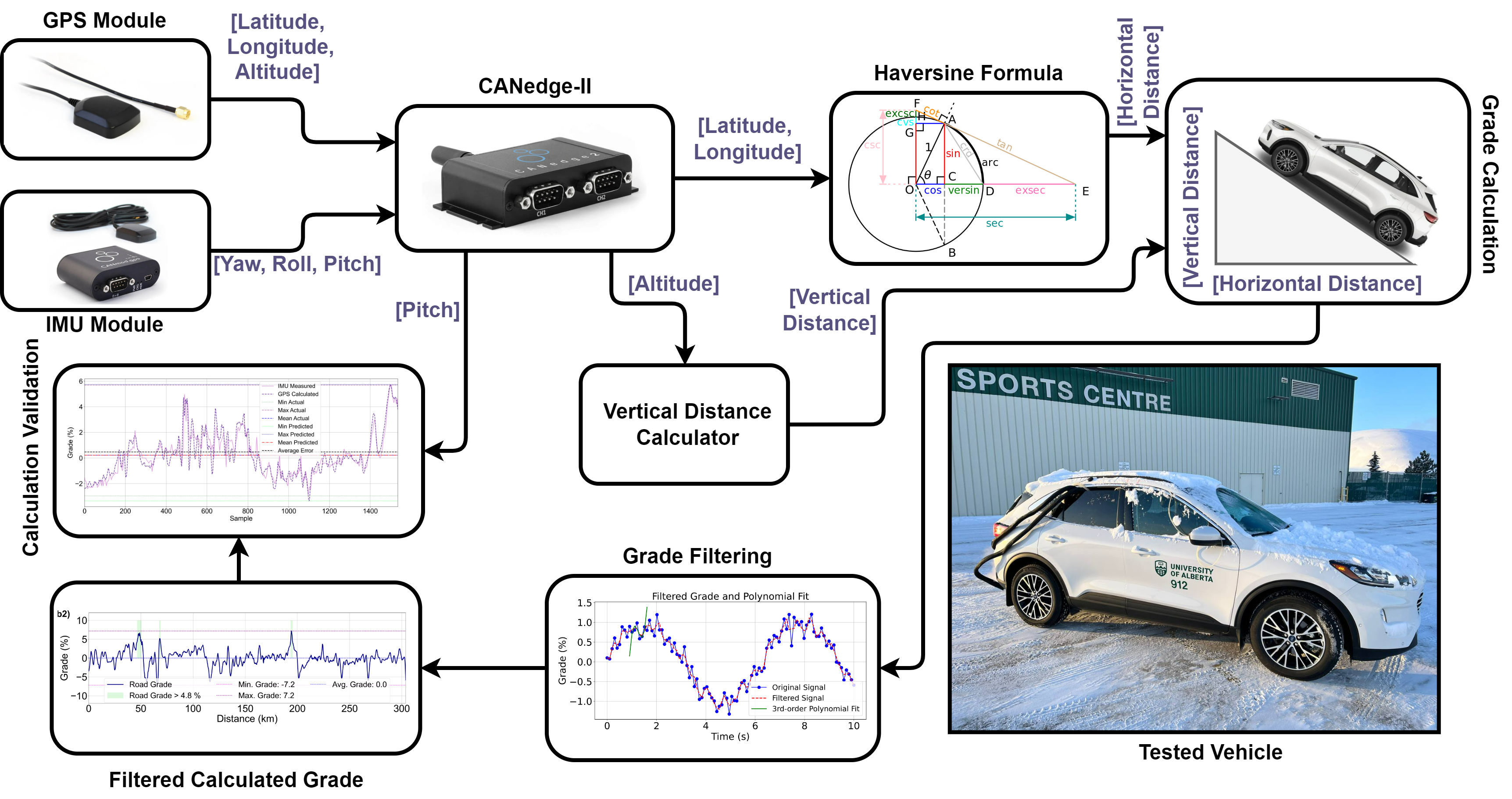}
  \caption{Road Grade Calculation Workflow and the Tested Vehicle.}
  \label{fig:road_grade}
\end{figure}

The methodology to calculate the road grade for various tests using GPS data obtained from CAN signals is described. The data was collected through the OBD port using the CANedge-II device from CSS Electronics. 
The road grade calculation follows the systematic process shown in Figure~\ref{fig:road_grade}. First, GPS data, including latitude, longitude, and altitude, is obtained using the GPS module of the CANedge II device. The horizontal distance between consecutive data points is then calculated using the Haversine formula \cite{chopde2013landmark} as:
\begin{equation}\label{eq:haversine}
\begin{aligned}
h &= \sin^2\left(\frac{\Delta\phi}{2}\right) + \cos(\phi_1) \cdot \cos(\phi_2) \cdot \sin^2\left(\frac{\Delta\lambda}{2}\right), \\
d &= R \cdot 2 \cdot \arctan2\left(\sqrt{h}, \sqrt{1-h}\right), \\
\end{aligned}
\end{equation}

where \( d \) represents the horizontal distance between two sampled data points. The terms \( \phi_1 \) and \( \phi_2 \) denote the latitudes of the first and second data points, respectively. Additionally, \( \Delta \phi \) and \( \Delta \lambda \) correspond to the differences in latitude and longitude between the two points. Finally, \( R \) is the Earth's radius, which is considered to be 6,371 km in this study.

Simultaneously, the vertical distance is determined by computing the difference in elevation values from the GPS data. Using these values, the road grade is derived as follows:
\begin{equation}\label{eq:road_grade}
g_{raw} (\%) = 100 \cdot \frac{\Delta \text{Altitude}}{d}
\end{equation}

To reduce GPS noise, the raw grade value, \(g_{\text{raw}}\), Eq.~(\ref{eq:road_grade}), is filtered with a Savitzky-Golay filter \cite{gorry1990general}:
\begin{equation}\label{eq:savitzky-golay}
{[g_{\text{F}}]}_k = \sum_{i=0}^{n} c_i \cdot {[g_{\text{raw}}]}_{k+w-i},
\end{equation}
where \(g_{\text{raw}}\) represents the raw grade, \(g_{\text{F}}\) represents the filtered grade, and \(c_i\) are the filter coefficients. The window size \(w\) must be an odd integer, and the coefficients \(c_i\) are computed using least-squares methods \hl{where full details of methods are provided in} \cite{savitzky_golay_derivation}.

A sliding window of size \(w = 25\) and a polynomial of order \(n = 3\) were selected, iterating over the entire dataset to effectively smoothen the grade profile by reducing high-frequency noise. The value at the center of the window is then replaced with the corresponding value from the fitted polynomial.

\begin{figure}[htbp]
  \centering
  \setlength{\abovecaptionskip}{0pt} % Adjust space between caption and the figure
  \resizebox{0.65\textwidth}{!}{%
    \begin{tabular}{cc}
      {\includegraphics[width=0.49\textwidth]{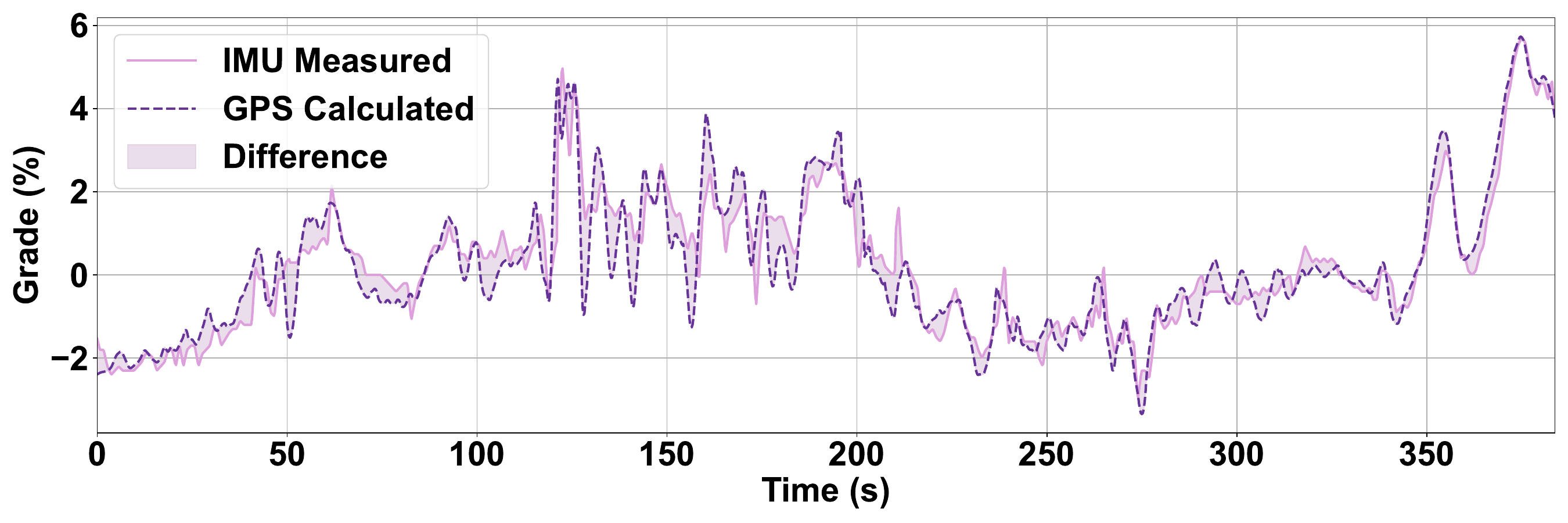}}\\ 
    \end{tabular}%
  }
  \caption{The calculated and measured grade vs time}
  \label{fig:road_grade_validation}
\end{figure}

The validation of the calculated road grade using GPS data is illustrated in Figure~\ref{fig:road_grade_validation}, including the calculated and measured grade using the Inertial Measurement Unit (IMU) device from CSS Electronics along the route shown in Figure~\ref{fig:grade_validation} and the error histogram of the grade calculation. The road grade calculation based on GPS data is validated by a Mean Absolute Error (MAE) of 0.47\%, a Mean Squared Error (MSE) of 0.41\%, and a Root Mean Squared Error (RMSE) of 0.64\%. \hl{The observed errors may arise from multiple factors, including the assumption of a fixed Earth radius during computations, inherent GPS measurement noise, and amplitude attenuation introduced by the applied Savitzky–Golay filter.}

\section{Representative Driving Cycle Construction Methods}
The methodology for constructing the proposed representative driving cycle is presented in this section and the benchmark approaches against which its performance is evaluated are outlined.

\subsection{Physics-Informed Expected SARSA-Monte Carlo (PIESMC)}\label{method}
A physics-informed expected SARSA and Monte Carlo (PIESMC) algorithm is used to generate drive cycles whose representativeness is defined by the closeness of their kinematic fragments to those in the experimental data. The PIESMC method offers improvements compared to the existing approaches in the literature to provide a robust, data-driven framework for generating representative driving cycles, which is both computationally efficient and more reflective of real-world driving behavior by having the following characteristics:
\begin{enumerate}
    \renewcommand{\labelenumi}{\roman{enumi}.}
    \item Efficient State-Action Learning: the use of the physics-informed expected SARSA algorithm constrains learning to feasible transitions and the intrinsic reward function, significantly reducing the computational burden.
    \item Realistic Idle Representation: the methodology explicitly incorporates idling conditions, ensuring that these critical segments are accurately represented within the generated driving cycle.
    \item Enhanced Drive Cycle Representativeness through Monte Carlo Integration: by integrating the Monte Carlo algorithm with SARSA, this method enables post-construction learning of driving cycles based on representativeness that is defined through a cost function derived from the kinematic fragments, ensuring the generated cycle reflects real-world driving conditions accurately.
\end{enumerate}

The proposed approach builds upon the MCB methodology for representative driving cycle construction and formulates the problem as a Markov decision process (MDP) to be solved using reinforcement learning (RL). In this context, an agent is defined as an autonomous decision-making entity that interacts with the environment by selecting actions based on a learned policy~\cite{sutton1998reinforcement}. To accelerate the process, the PIESMC method learns the optimal next state based on the reward received upon transitioning to that state from a specific starting state. The state is defined by speed, acceleration, and road grade. This enables the agent to identify a policy that maximizes rewards while navigating the large state-action space characteristic of driving cycle construction. Furthermore, the Monte Carlo method is used to enhance learning after the completion of each constructed driving cycle, as an episode. Once an episode concludes, all visited states are stored, and their values are updated using the Monte Carlo method. This process enables the agent to learn not only from the representativeness of individual transitions but also from an evaluation of the overall driving cycle's representativeness at the end of each episode. This feature ensures that the constructed driving cycle closely aligns with real-world driving conditions.

A significant challenge for most reinforcement learning (RL) algorithms, such as SARSA, is managing large state and action spaces, which often demand extensive interactions with the environment and result in a computationally intensive learning process \cite{andriotis2019managing}. To address this, the proposed method introduces an optimized approach where the agent identifies feasible state transitions for each visited state and focuses exclusively on those transitions. Using knowledge derived from experimental data, this strategy enables more efficient learning while significantly reducing computational demands. Another technique implemented in this study to enhance learning efficiency is the use of \hl{a count-based intrinsic reward function}~\cite{tang2017exploration}. This approach encourages the agent to explore more extensively, facilitating faster convergence to optimal actions.

\begin{figure}[htbp]
  \centering
  \includegraphics[width=0.80\textwidth]{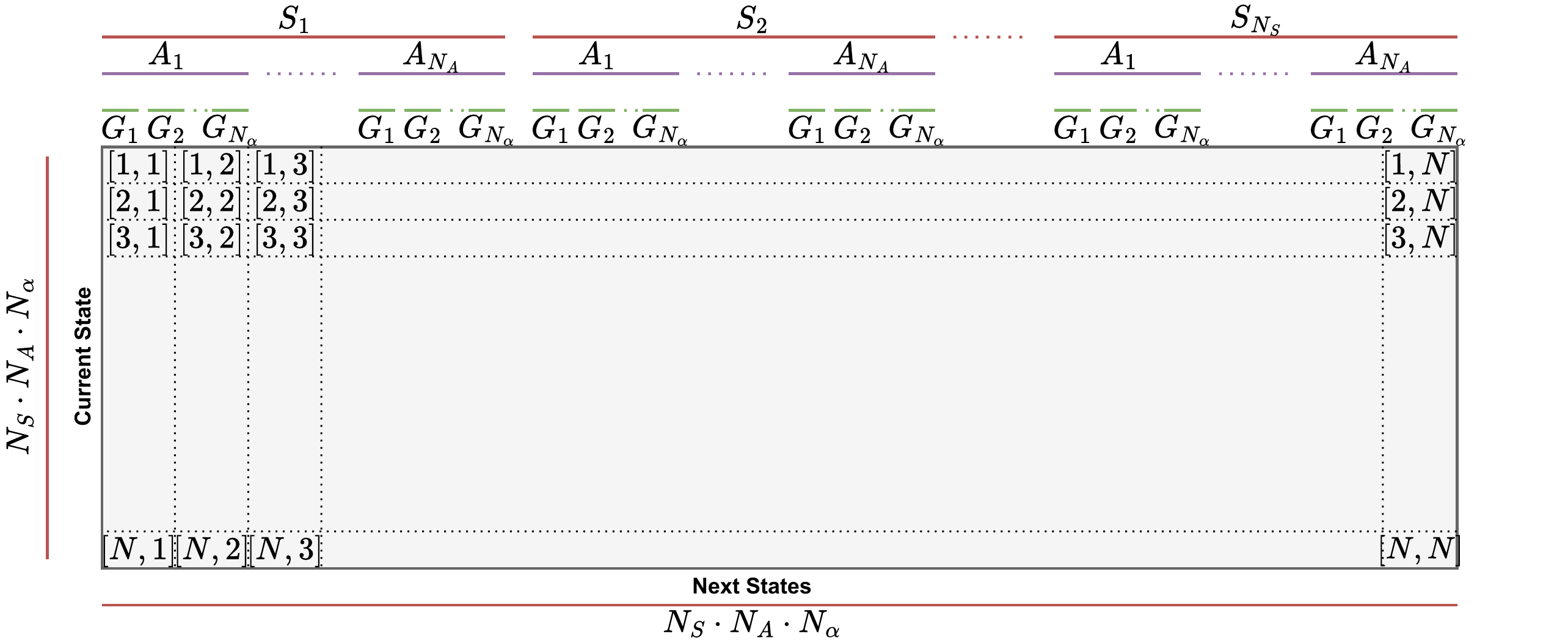}
  \caption{SAGSTM structure. \(S_i\) is the speed bins \(i^{th}\), \(A_j\) is the \(j^{th}\) acceleration bin, and \(G_k\) is the \(k^{th}\) grade bin. \(N_S\) is the number of speed bins, \(N_A\) is the number of acceleration bins, and \(N_{\alpha}\) is the number of road grade bins.}
  \label{fig:SAGSTM}
\end{figure}

The Speed Acceleration Grade State Transition Matrix (SAGSTM) is an essential tool for addressing the problem of representative driving cycle construction within the framework of the Markov chain process. The SAGSTM is constructed using the available experimental data. Figure~\ref{fig:SAGSTM} illustrates the structure of the SAGSTM, while Eq.~(\ref{eq:SAGST}) defines the calculation process of each cell within the matrix. 

\noindent By defining the state space as:
\begin{equation}\label{eq:state_space}
  \mathcal{S} = \{\, s_{ijk} : i = 1, \dots, N_S;\; j = 1, \dots, N_A;\; k = 1, \dots, N_{\alpha} \,\},
\end{equation}

\noindent where \(N_S\) is the number of speed bins, \(N_A\) is the number of acceleration bins, and \(N_{\alpha}\) is the number of grade bins. The \(\mathbf{\text{SAGSTM}} \in \mathbb{R}^{N \times N}\) is calculated as: 

\begin{equation}\label{eq:SAGST}
\text{SAGSTM}(l,m) = P(s_m | s_l) = \frac{\sum_{d=1}^{D}N(s_l \rightarrow s_m)}{\sum_{d=1}^{D}\sum_{k=1}^{N} N(s_l \rightarrow s_k)}, \quad l, m \in \{1, 2, \dots, N\},
\end{equation}

\noindent where the total number of states is \( N = N_S \times N_A \times N_{\alpha}\) and \(D\) denotes the total number of datasets used in this study. For each dataset \(d \in \{1,\dots, D\}\), let \(N(s_l \rightarrow s_m)\) denote the number of transitions from state \(s_l\) to state \(s_m\). Moreover, this definition satisfies the normalization conditions:
\begin{equation}
  \sum_{m=1}^{N} P(s_m \mid s_l) = 1 \quad \text{and} \quad P(s_m \mid s_l) \geq 0, \quad \forall\, l=1,\dots,N, 
\end{equation}

However, as many transitions are not observed in the experimental data, most of the entries in the SAGSTM are zero. Let \(\{e_1, e_2, \dots, e_N\}\) denote the standard basis vectors in \(\mathbb{R}^N\). Then, the SAGSTM can be expressed in sparse form as:
\begin{equation}
  \text{SAGSTM} = \sum_{(l,m) \in \mathcal{I}} P(s_m \mid s_l) \, e_l e_m^{\top},
\end{equation}
where the set of feasible states will be defined as:
\begin{equation}
  \mathcal{I} = \{ (l,m) \in \{1,\dots,N\}^2 : P(s_m \mid s_l) \neq 0 \}.
\end{equation}
Since \(|\mathcal{I}| \ll N^2\), the matrix SAGSTM is sparse. This sparsity facilitates efficient storage and computation and allows the proposed method to scale to larger state spaces, thereby including road-grade transitions.
 
The discounted reward (or return) that the agent in the proposed method tends to maximize its expected value will be defined as given:

\begin{equation}\label{eq:return}
G_q = \sum_{p=0}^\infty \gamma^p R_{q+p}, 
\end{equation}
where \( G_q \) is the total discounted reward starting at time \( q \). Moreover, \( R_q, R_{q+1}, R_{q+2}, \dots \) denote the rewards received by the agent at consecutive time steps \( q, q+1, q+2, \dots \) and \( \gamma \) is the discount factor, which is a value between 0 and 1 and determines the importance of future rewards. The term \( \gamma^p \) applies a discount to the reward at time step \( q+p \), progressively reducing the weight of rewards as \( p \) increases.

The construction of the representative driving cycle using the Expected SARSA-Monte Carlo methodology is formulated as an MDP, with its components being defined as follows.
The state set \( \mathcal{S} \) consists of integers ranging from 1 to \( N \), where \( N \) represents the total number of unique combinations of speed bins \( N_S \), acceleration bins \( N_A \), and road grade bins \( N_{\alpha} \). \noindent Each state corresponds to a specific combination of speed, acceleration, and road grade, which is expressed as:
\begin{equation}\label{eq:state_space}
\mathcal{S} = \left\{ s_{n} \in \mathbb{N} \;:\; s_{n} = n,\quad n=1,\dots,N \right\}, 
\end{equation}

\noindent each state index \(n\) is uniquely determined by the triple \((i,j,k)\) via the mapping:
\begin{equation}\label{eq:L_definition}
n = (i-1)(N_A \cdot N_{\alpha}) + (j-1)N_{\alpha} + k,\quad i = 1, \dots, N_S,\quad j = 1, \dots, N_A,\quad k = 1, \dots, N_{\alpha}.
\end{equation}
For each state, the action set \( \mathcal{A}(s) \) is defined as being identical to the state set \( \mathcal{S}\), allowing transitions between all possible states as:
\begin{equation}\label{eq:action_space}
\mathcal{A}(s) = \mathcal{S}, \quad \forall s \in \mathcal{S}.
\end{equation}

\noindent The action space for the agent could be excessively large, which can hinder the efficient learning process. To address this problem, the physics of the problem is leveraged to accelerate the expected SARSA algorithm by eliminating infeasible actions based on \hl{physical constraints derived from experimental data.} Specifically, transitions involving significant changes in speed, acceleration, and grade that are unrealistic or absent in the experimental dataset are excluded. The proposed algorithm identifies feasible transitions using the SAGSTM, where zero values indicate impossible transitions. At each state, the agent considers the SAGSTM and ignores transitions corresponding to zero values, ensuring the learning process is based on feasible scenarios. The representation of the feasible action set is defined as:

\begin{equation}
\mathcal{A}'(s) = \{s' \in \mathcal{S} \mid \text{SAGSTM}(s, s') \neq 0\}.
\end{equation}

The extrinsic reward function $R_{ES}^{ext}: \mathcal{S} \times \mathcal{A'} \to \mathbb{R}$ for expected SARSA learning is defined using the value of the SAGSTM which represents the probability of transitioning from state s to state a. To ensure the reward encourages likely transitions, the reward function is modeled as a softmax function of \( \text{SAGSTM}(s, a) \) as:
\begin{equation}\label{eq:ext_reward_function}
R_{ES}^{ext}(s, a) =
\frac{\exp(\tau\cdot\text{SAGSTM}(s, a))}{\sum_{a' \in \mathcal{A}(s)} \exp(\tau\cdot\text{SAGSTM}(s, a'))},
\end{equation}

where the parameter \(\tau\) acts as a factor that determines the sensitivity of the reward function to differences in the values of \(\text{SAGSTM}(s,a)\). A larger \(\tau\) amplifies these differences, encouraging the agent to strongly favor the action with the highest reward distribution. Conversely, a smaller \(\tau\) smooths out the differences, resulting in a more uniform reward landscape that diminishes the agent's incentive to select the highest-valued action.

To further encourage the agent to explore, an intrinsic reward function $R_{ES}^{int}: \mathcal{S} \times \mathcal{A'} \to \mathbb{R}$ is employed, which uses a count-based method as:

\begin{equation}\label{eq:int_reward}
R_{ES}^{int}[s,a] = \frac{\beta}{\sqrt{\mathcal{M}[s,a]}},
\end{equation}

in which $\mathcal{M}(s,a)$ is the number of times the state $s$ has been visited and the action $a$ has been chosen, and $\beta$ is a scaling factor. This method tracks the number of times each state is visited and provides an incentive to the agent to explore less-visited states. Such exploration accelerates convergence to optimal actions \cite{tang2017exploration}:

The final reward, $R(s,a)$, is then computed as the sum of the extrinsic reward $R^{ext}(s,a)$ and the intrinsic reward $R^{int}(s,a)$, given by:
\begin{equation}
R_{ES}[s,a] = \lambda_{ext} R^{ext}[s,a] + \lambda_{int} R^{int}[s,a].
\end{equation}

Both the expected SARSA algorithm and the Monte Carlo method are employed to optimize the agent's learning process. The reward function in the Monte Carlo method is based on the cost function, which is a function of the error in kinematic fragments as defined in Table~\ref{tab:kin_def}.

\begin{table}[htbp]
\centering
\caption{Mathematical definitions and descriptions of kinematic fragments.}
\label{tab:kin_def}
\resizebox{\textwidth}{!}{%
  \setlength{\tabcolsep}{8pt}% adjust horizontal padding
  \renewcommand{\arraystretch}{1.0}% adjust vertical spacing between rows
  \begin{tabular}{@{}ll ll@{}}
    \toprule
    \textbf{Symbol} & \textbf{Description} & \textbf{Symbol} & \textbf{Description} \\
    \midrule
    \(\bar{V}_{EI}\) & Average speed computed for \(V_i>0.025\,\mathrm{m/s}\) (excludes idling) 
      & \(t_i\) & Percentage of time idling (\(V\le0.025\,\mathrm{m/s}\)) \\[2ex]
    \(\bar{V}\) & Overall average speed across the cycle 
      & \(t_c\) & Percentage of time cruising (\(V>5\,\mathrm{m/s}\) and \(-0.15\le A\le0.15\,\mathrm{m/s}^2\)) \\[2ex]
    \(\bar{A}_p\) & Average positive acceleration (\(A_i\ge0.15\,\mathrm{m/s}^2\)) 
      & \(t_{ap}\) & Percentage of time with positive acceleration (\(A>0\)) \\[2ex]
    \(\bar{A}_n\) & Average negative acceleration (\(A_i\le-0.15\,\mathrm{m/s}^2\)) 
      & \(t_{an}\) & Percentage of time with negative acceleration (\(A<0\)) \\
    \bottomrule
  \end{tabular}%
}
\end{table}

% \begin{enumerate}
%     \renewcommand{\labelenumi}{\roman{enumi}.}
%     \item \(\bar{V}_{EI}\): the average speed computed from data points where the vehicle speed exceeds 0.025\,m/s (thus excluding idling).
%     \item \(\bar{V}\): the overall average speed across the entire cycle.
%     \item \(\bar{A}_p\) and \(\bar{A}_n\): the average positive and negative accelerations, respectively, computed from acceleration values exceeding 0.15\,m/s\(^2\) in magnitude.
%     \item \(t_i\): the percentage of time during which the vehicle is idling (speed \(\leq\) 0.025\,m/s).
%     \item \(t_c\): the percentage of time during which the vehicle is cruising (characterized by a speed greater than 5\,m/s and acceleration between -0.15\,m/s\(^2\) and 0.15\,m/s\(^2\)).
%     \item \(t_{ap}\) and \(t_{an}\): the percentages of time during which the vehicle experiences positive and negative acceleration, respectively.
% \end{enumerate}

For each kinematic fragment \(i\) (with \(i = 1, \dots, N_f\)), an error \(\epsilon_i\) is computed that quantifies the deviation between the observed and the reference values based on the experimental data for that fragment with the overall cost function is defined as:
\begin{equation}\label{eq:cost_function}
E = \sum_{i=1}^{N_f} \epsilon_i.
\end{equation}

\noindent The reward function for the Monte Carlo method is given by:
\begin{equation}\label{eq:mc_reward}
R_{MC} = \frac{\varsigma}{1 + E},
\end{equation}
where \(\varsigma\) is a scaling factor that controls the magnitude of the reward. This ensures that lower cumulative errors in the kinematic fragments (i.e., lower values of \(E\)) yield higher rewards, encouraging the agent to select actions that minimize the overall cost.

Lastly, the dynamics of the environment are deterministic $\mathcal{P}: \mathcal{S} \times \mathcal{A} \times \mathcal{S} \to [0,1]$, meaning that the probability of transitioning from state \( s \) to state \( s' \) after taking action \( a \) is defined in:
 
\begin{equation}\label{eq:env_dyn}
\mathcal{P}(s' | s, a) = 
\begin{cases} 
1, & \text{if the transition from } s \text{ to } s' \text{ via } a \text{ is valid} \\
0, & \text{otherwise}
\end{cases}.
\end{equation}

The expected SARSA algorithm works based on the estimation of state value functions \(V^{\pi}(s)\), which is defined as the value of expected return \(\mathbb{E_{\pi}}[G_t]\) when the agent is in the state \(s\) and follows the policy $\pi : \mathcal{S} \to \mathcal{A'}$, the same as all the temporal difference (TD) algorithms \cite{tesauro1995temporal}. Furthermore, the action-values function is defined as the expected return when the agent takes action \(a\) in state \(s\), and then follows the policy \(\pi\). So, the state value function is:
\begin{equation}\label{eq:state_value}
V_{\pi}[s] = \sum_{a \in \mathcal{A}'(s)} \pi(a \mid s) \cdot Q[s, a].
\end{equation}

The iterative update rule of the state-action values based on the TD(0) methods is given in: 
\begin{equation}\label{eq:td0_update}
Q_{n+1}[s, a] = Q_n[s, a] + \alpha \cdot \delta_n, \quad \text{where} \quad s \in \mathcal{S}, \, a \in \mathcal{A}'(s),
\end{equation}

where \(n\) denotes the iteration index, and the TD(0) error, \(\delta_n\),  for the Q-learning method is given by:
\begin{equation}\label{eq:delta_q_learning}
\delta_n = R_{ES}[s,a] + \gamma \cdot max_{\pi}~Q_n[s',a']\ - Q_n[s, a].
\end{equation}

In Eq.~(\ref{eq:delta_q_learning}), the \( \max \) operator ensures that the estimation policy is greedy, resulting in a deviation from the actual policy being followed. This deviation classifies Q-learning as an off-policy algorithm. The optimal state-action values \(Q^*[s, a]\) can be determined through an iterative update of the state-action values \(Q[s, a]\). Once \(Q^*[s, a]\) is obtained, the optimal policy \(\pi^*\) can be derived directly from these values. Additionally, the \( \max \) operator guarantees that \( Q[s, a] \) converges to the optimal action-value function \( Q^*[s, a] \).
In contrast, the SARSA algorithm maintains the same behavior and estimation policies, making it an on-policy algorithm. Consequently, the update rule for the SARSA algorithm is expressed as shown in:
\begin{equation}\label{eq:delta_sarsa}
\delta_n = R_{ES}[s,a] + \gamma \cdot Q_n[s',a'] - Q_n[s, a].
\end{equation}
\hl{Since SARSA is an on-policy reinforcement learning algorithm, its convergence is contingent upon all states being visited infinitely often, necessitating adequate exploration. Typically, in both Q-learning and SARSA, a stochastic behavior policy is employed to ensure sufficient exploration for convergence. However, this inherent stochasticity introduces variance into the update process. To mitigate this variance, the learning rate $\gamma$ must typically be reduced, which subsequently slows the overall learning progression. Expected SARSA, a variation of the SARSA algorithm implemented in this study, specifically addresses this challenge by reducing the variance associated with updates. By decreasing update variance, Expected SARSA facilitates the use of a higher learning rate $\gamma$, thereby accelerating the learning process and achieving faster convergence. The TD(0) error, $\delta$, for the Expected SARSA algorithm is defined as follows:}

\begin{equation}\label{eq:delta_expected_sarsa}
\delta_n = R_{ES}[s,a] + \gamma \cdot \mathbb{E}_{\pi}\{Q_n[s',a']\} - Q_n[s, a]
\end{equation}

The expected state value will be calculated as:

\begin{equation}\label{eq:expected_state_value}
\mathbb{E}_{\pi}\{Q_n[s',a']\} = \sum_{a' \in \mathcal{A}'(s')} \pi(a' \mid s') \cdot Q_n[s', a']
\end{equation}
where \(\mathcal{A}'(s') \subseteq \mathcal{A}(s')\) represents the feasible action set for the state \(s'\), as determined by the physics of the problem and the available experimental data. While the conventional calculation of the expected value sums over all possible actions in \(\mathcal{A}(s')\), this formulation restricts the summation to \(\mathcal{A}'(s')\), ensuring that only feasible actions contribute to the expected state value.
This modification aligns the learning process with the constraints of the physical system and experimental data. 

Given that there are two separate reward functions for expected SARSA and Monte Carlo methods, having one state action value for both methods can cause instability in the updates. Therefore, a different state-action value will be used for the Monte Carlo method. Once the episode terminal is reached and the Monte Carlo state-action values are updated. The update rule for the state-action value function $Q[s, a]$ using the Monte Carlo method is defined as:
\begin{equation}\label{eq:mc_update}
Q^{MC}_{n+1}[s, a] \gets Q^{MC}_n[s, a] + \alpha \left( G^{MC}_q - Q^{MC}_n[s, a] \right),
\end{equation}
where $G^{MC}_q$ represents the return starting from time $q$ and will be calculated using Eq. (\ref{eq:return}) and Eq. (\ref{eq:mc_reward}) as:
\begin{equation}\label{eq:mc_return}
G_q = \gamma_{MC}^{T-q} R_{MC},
\end{equation}
where, $T \in \mathbb{N}$ is the number of steps required to reach the terminal state or the duration of the driving cycle, $R_{MC}$ represents the Monte Carlo reward function based on the representativeness of the driving cycle, and $\gamma_{MC}$ is the discount factor for the Monte Carlo Method and $\alpha$ is the learning rate. Depending on the approach, this update can be applied to the first visit or every visit of the state-action pair $(s, a)$ during an episode.

\begin{figure}[htbp]
  \centering
  \includegraphics[width=0.95\textwidth]{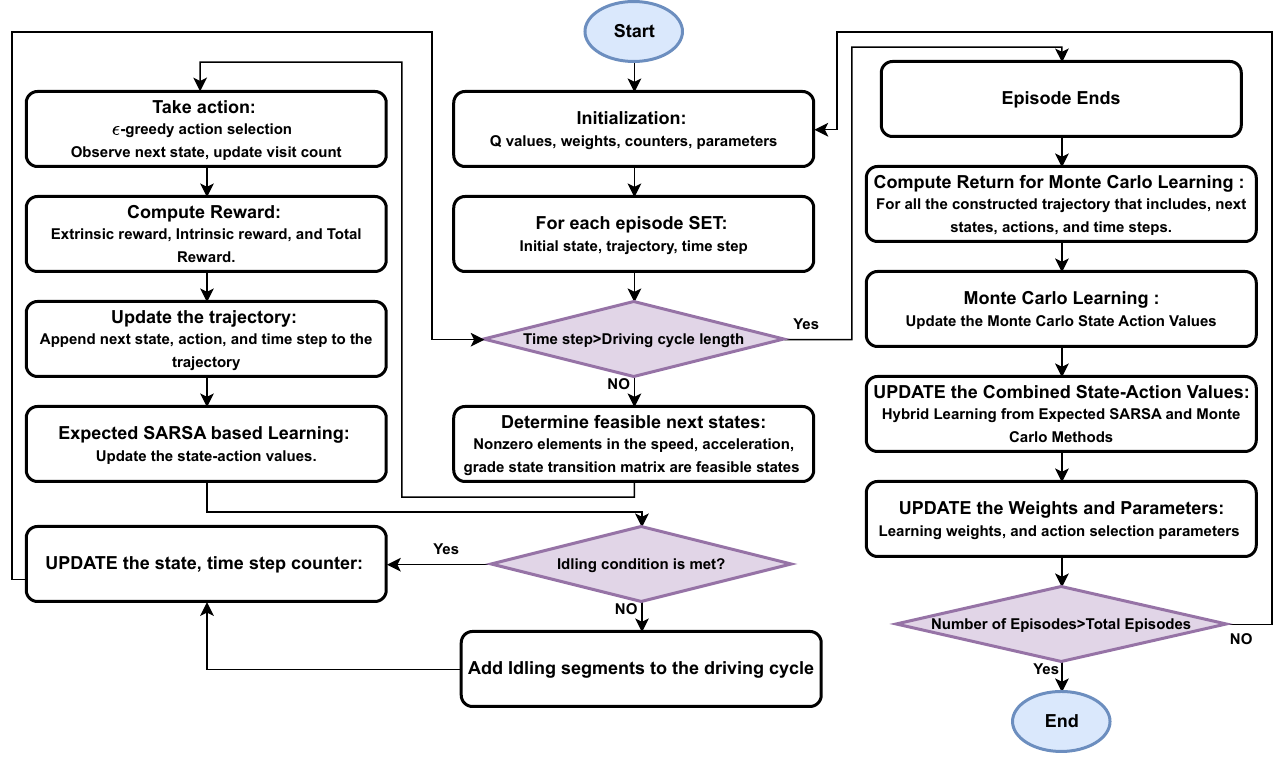}
  \caption{The Proposed Physics Informed Expected SARSA and Monte Carlo Method Workflow}
  \label{fig:piesmc_workflow_nv}
\end{figure}

Finally, the combined state action values will be a weighted sum of state-action values from expected SARSA and Monte Carlo methods as:
\begin{equation}\label{eq:combined_q}
Q_{Combined}[s, a] = w_{MC}Q_{MC}[s, a] + w_{ES}Q[s, a],
\end{equation}

with \hl{the proposed methodology workflow shown in Figure}~\ref{fig:piesmc_workflow_nv} summarized in Algorithm~\ref{algo:expected_sarsa_mc}.

\begin{algorithm}[ht!]
\small
\SetAlgoLined
\SetKwInOut{Input}{Input}
\SetKwInOut{Output}{Output}
\Input{State-action space \( \mathcal{S}, \mathcal{A} \), discount factors \( \gamma_{ES}, \gamma \), learning rates \( \alpha_{ES}, \alpha \), initial \( \epsilon \), minimum \( \epsilon_{\text{min}} \), decay rate, weights \( \lambda_{ext}, \lambda_{int} \), parameters \( \beta \), \( \tau \), initial weight \( w_{ES,0} \), minimum weight \( w_{ES,\text{min}} \), and the SAGSTM.}
\Output{Updated \( Q_{Combined}[s, a] \) and policy \( \pi \).}

\textbf{Initialization:} Initialize \( Q_{ES}[s, a] \), \( Q_{MC}[s, a] \), and \( Q_{Combined}[s, a] \) for all \( s \in \mathcal{S}, a \in \mathcal{A} \). Initialize \( Q_{ES}[s, a] \) using SAGSTM. Initialize visit count \( \mathcal{M}[s,a] \gets 0 \) for all \( s \in \mathcal{S}, a \in \mathcal{A} \). Set \( w_{ES} \gets w_{ES,0} \) and \( w_{MC} \gets 1 - w_{ES} \). \\

\ForEach{episode}{
    Initialize state \( s \) to one of the initial states in the experimental data. \\
    Initialize an empty list \( \mathcal{T} \) to store the trajectory of state-action pairs and time steps. \\
    Set time step \( q \gets 0 \). \\

    \While{\textit{not terminal state}}{
        Determine feasible actions \( \mathcal{A}'(s) \) based on SAGSTM. \\
        Select action \( a \) using adaptive \(\epsilon\)-greedy policy: \\
        \Indp With probability \( \epsilon \), choose a random action. \\
        Otherwise, choose \( a = \text{argmax}_{a' \in \mathcal{A}'(s)} Q_{Combined}[s, a'] \). \\
        \Indm
        Take action \( a \), observe transition to next state \( s' \). \\
        Update visit count \( \mathcal{M}[s,a] \gets \mathcal{M}[s,a] + 1 \). \\
        Compute extrinsic reward: \\
        \(
        R^{ext}_{ES}(s, a) = 
        \frac{\exp(\tau \cdot \text{SAGSTM}(s, a))}{\sum_{a' \in \mathcal{A}'(s)} \exp(\tau \cdot \text{SAGSTM}(s, a'))}.
        \)\\
        Compute intrinsic reward: \( R^{int}_{ES}(s,a) = \frac{\beta}{\sqrt{\mathcal{M}[s,a]}} \). \\
        Compute total reward: \( R_{ES}[s,a] = \lambda_{ext} R^{ext}_{ES}[s,a] + \lambda_{int} R^{int}_{ES}[s,a] \). \\
        Append \( (s, a, q) \) to \( \mathcal{T} \). \\
        Compute \( V_{\pi}[s'] = \sum_{a' \in \mathcal{A}(s')} \pi(a'|s') \cdot Q_{ES}[s', a'] \). \\
        Update \( Q_{ES}[s, a] \) using Expected SARSA update: \\
        \( Q_{ES}[s, a] \gets Q_{ES}[s, a] + \alpha_{ES} \cdot [R_{ES} + \gamma_{ES} V_{\pi}[s'] - Q_{ES}[s,a]] \). \\
        Check the idling condition. If true, add an idling segment to the driving cycle. \\
        Set \( s \gets s' \). \\
        Increment \( q \gets q + 1 \).
    }
    
    Compute returns for all visited state-action pairs in the trajectory: \\
    \ForEach{\((s, a, t) \in \mathcal{T}\)}{
        Compute \( G^{MC}_q = \gamma^{T-q} R_{MC} \), where \( T \) is the final time step of the episode. \\
        Update \( Q_{MC}[s, a] \): \\
        \( Q_{MC}[s, a] \gets Q_{MC}[s, a] + \alpha_{MC} \cdot [G^{MC}_q - Q_{MC}[s, a]] \).
    }
    Update \( Q_{Combined}[s, a] \): \\
    \( Q_{Combined}[s, a] \gets w_{ES} Q_{ES}[s, a] + w_{MC} Q_{MC}[s, a] \). \\
    
    Adaptively update \( w_{ES} \) and \( w_{MC} \): \\
    \( w_{ES} \gets \max(w_{ES,\text{min}}, w_{ES} \cdot \text{decay\_rate}) \). \\
    \( w_{MC} \gets 1 - w_{ES} \). \\
    
    Adaptively update exploration probability \( \epsilon \): \\
    \( \epsilon \gets \max(\epsilon_{\text{min}}, \epsilon \cdot \text{decay\_rate}) \).
}
\caption{The Proposed Physics Informed Expected SARSA and Monte Carlo Method.}
\label{algo:expected_sarsa_mc}
\end{algorithm}

\subsection{Microtrip-Based (MTB) Method}
\label{subsec:microtrip_method}

\begin{figure}[ht!]
  \centering
  \includegraphics[width=0.99\textwidth]{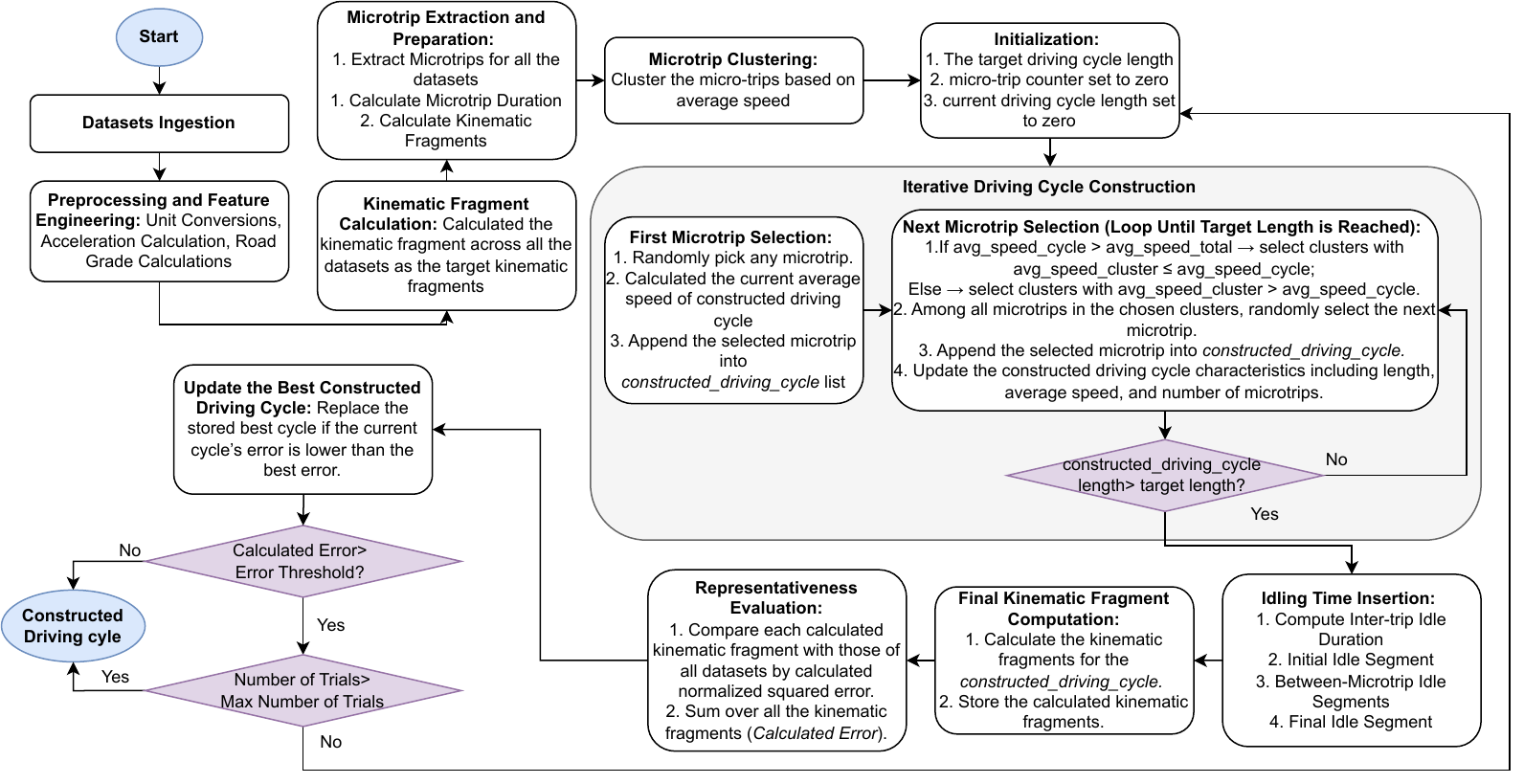}
  \caption{Microtrip‐based workflow for representative driving cycle construction}
  \label{fig:mtb_workflow}
\end{figure}

The microtrip‐based method divides a vehicle speed profile into ``microtrips" defined as segments between stops, characterizes each by kinematic metrics defined in Table\ref{tab:kin_def}, and then iteratively selects segments to reach a target duration $T_{\text{target}}$, drawing additional microtrips from clusters whose mean speeds align with the current cycle’s average, with idle intervals inserted based on original data idling average. Finally, kinematic metrics are computed and compared to original data kinematic targets via normalized squared errors. A detailed workflow is shown in Figure~\ref{fig:mtb_workflow}.

\subsection{Markov-Chain-Based (MCB) Method}

\begin{figure}[ht!]
  \centering
  \includegraphics[width=0.99\textwidth]{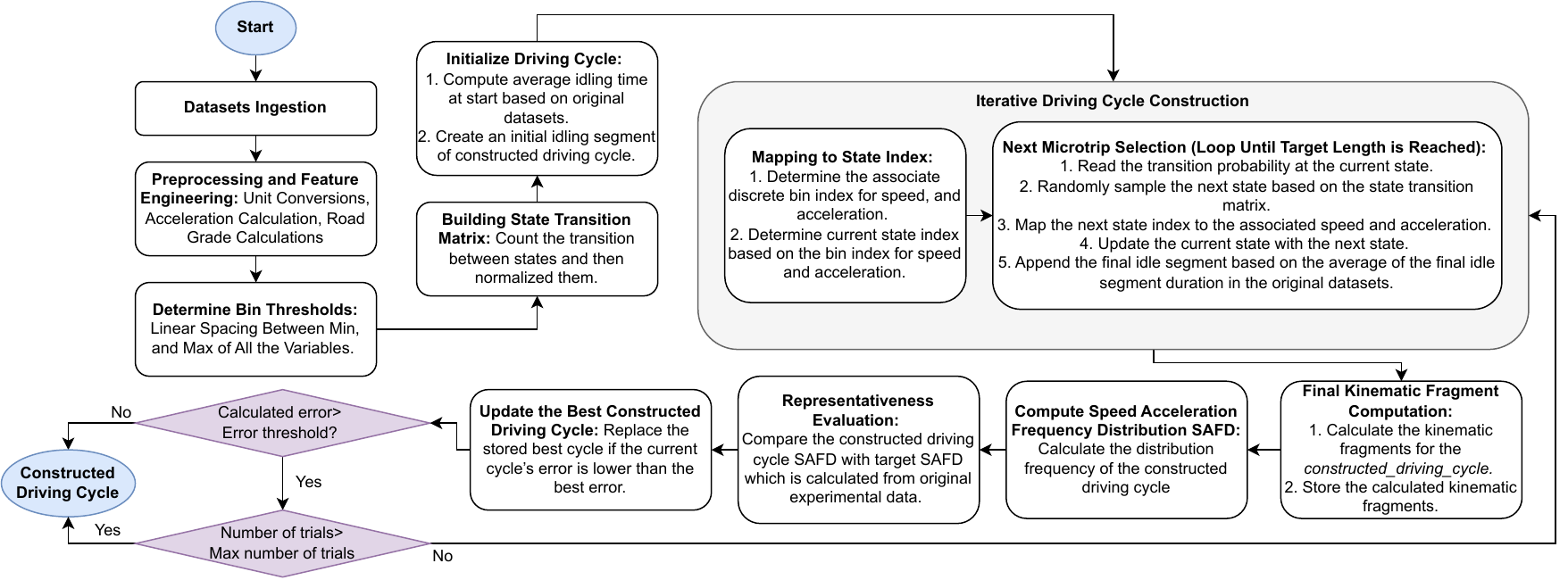}
  \caption{Markov-chain‐based workflow for representative driving cycle construction}
  \label{fig:mcw_workflow}
\end{figure}

The Markov‐chain‐based method constructs a representative driving cycle by first quantizing observed speed, acceleration, and grade values into discrete bins and then deriving a three‐dimensional Speed–Acceleration–Grade State Transition Matrix (SAGSTM) from the historical micro‐level traces. Beginning with an initial idle state (zero speed, zero acceleration, zero grade), consecutive states are sampled according to the conditional transition probabilities in the SAGSTM, thereby generating a sequence of (speed, acceleration, grade) tuples whose length equals the target cycle duration minus the average idle intervals at the start and end. After appending constant‐grade idle segments at both ends, the concatenated sequence is re‐indexed to ensure uniform time steps, and its kinematic fragments are computed. A multi‐bin Speed–Acceleration–Grade Frequency Distribution (SAGFD) is then calculated for the generated cycle and compared against the fleet‐level reference SAGFD; the total normalized error (i.e., the sum of squared differences across all bins) serves as the optimization objective. Detailed implementation steps, including bin threshold determination, idle insertion, and error computation, are illustrated in the workflow shown in Figure~\ref{fig:mcw_workflow}.

\section{Results and Discussions}\label{results}
The methodologies were implemented and evaluated on a computer equipped with a 12th Gen Intel\textsuperscript{®} Core\textsuperscript{™} i9-12700K CPU and Intel\textsuperscript{®} UHD 770 Graphics. The codes were executed using Python version 3.11.

\subsection{The proposed method's performance for driving cycle construction}\label{subsec:proposed-PIRL-based}
\begin{figure*}[ht!]
  \centering
  \setlength{\abovecaptionskip}{0pt} % Adjust space between caption and the figure
  \begin{subfigure}[t]{0.45\textwidth}
    \centering
    \caption{PIESMC DC 1}
    \label{fig:PIESMC_DC1}
    \includegraphics[width=\textwidth]{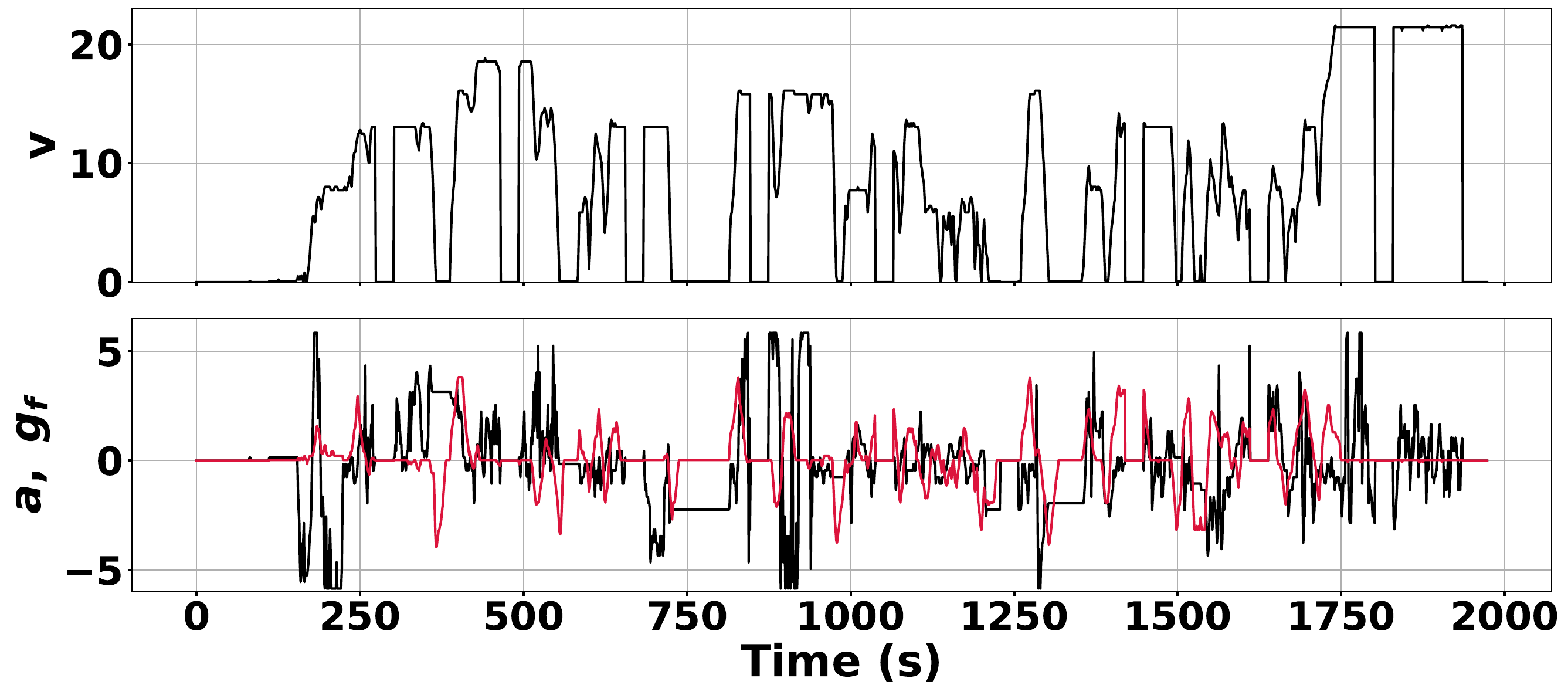}
  \end{subfigure} 
  \hspace{0.5em}
  \begin{subfigure}[t]{0.45\textwidth}
    \centering
    \caption{PIESMC DC 2}
    \label{fig:PIESMC_DC2}
    \includegraphics[width=\textwidth]{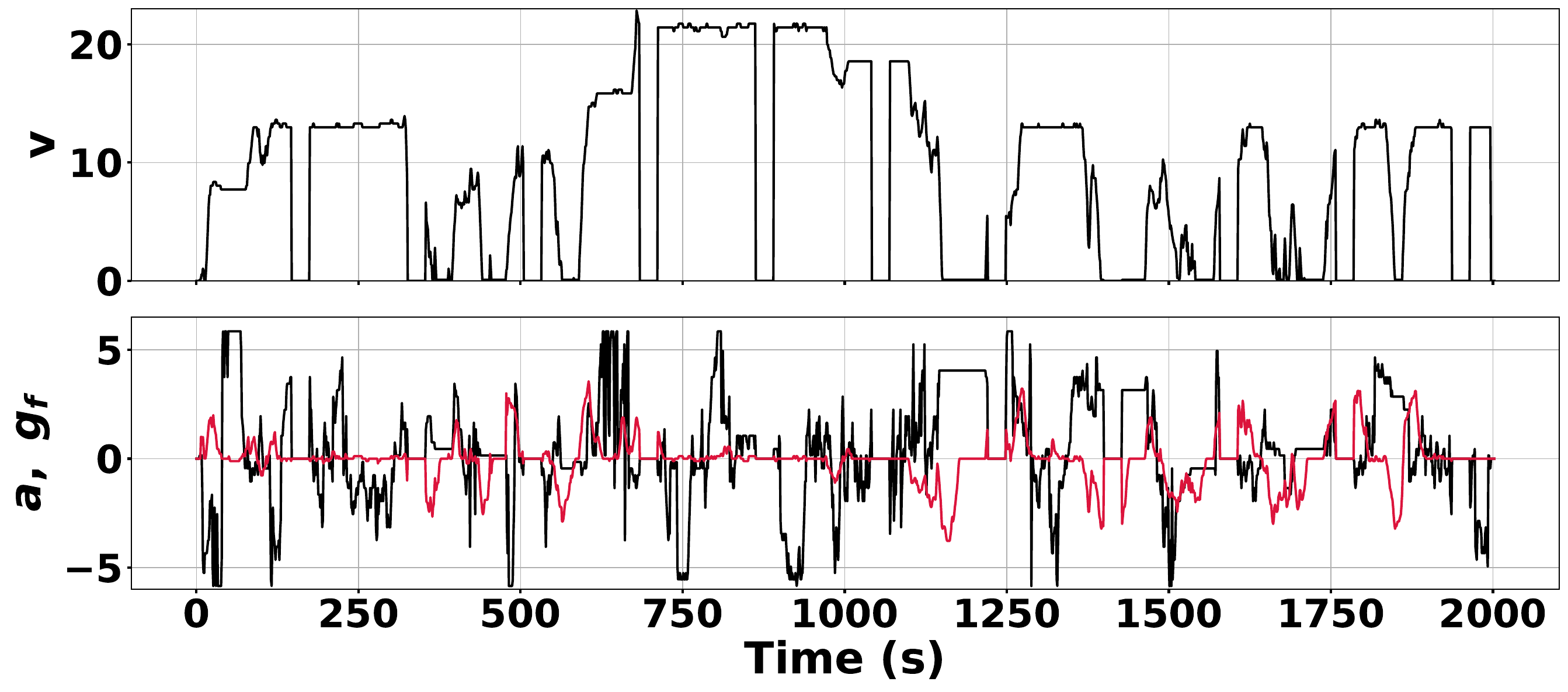}
  \end{subfigure}
  
  \vspace{0.5em}
  \begin{subfigure}[t]{0.45\textwidth}
    \centering
    \caption{MCB DC 1}
    \label{fig:MCB_DC1}
    \includegraphics[width=\textwidth]{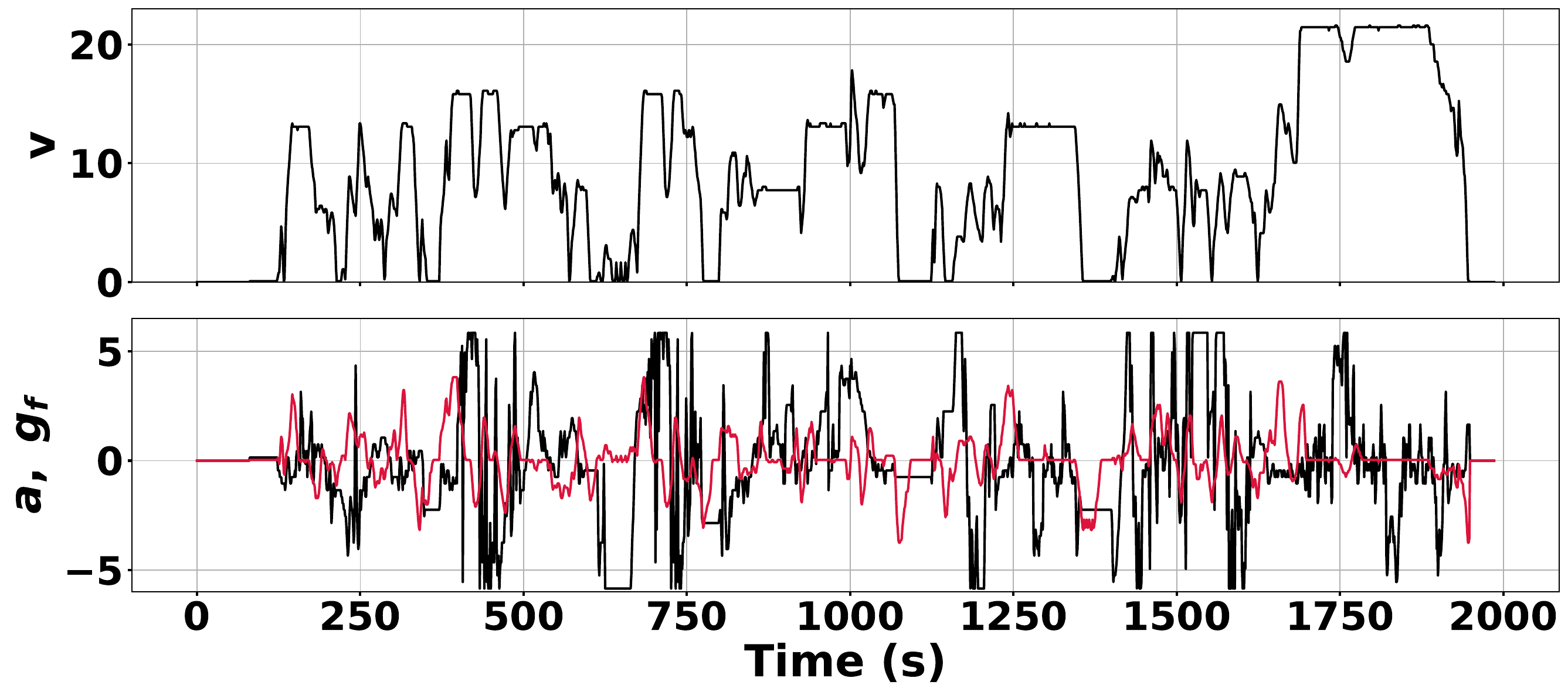}
  \end{subfigure} 
  \hspace{0.5em}
  \begin{subfigure}[t]{0.45\textwidth}
    \centering
    \caption{MCB DC 2}
    \label{fig:MCB_DC2}
    \includegraphics[width=\textwidth]{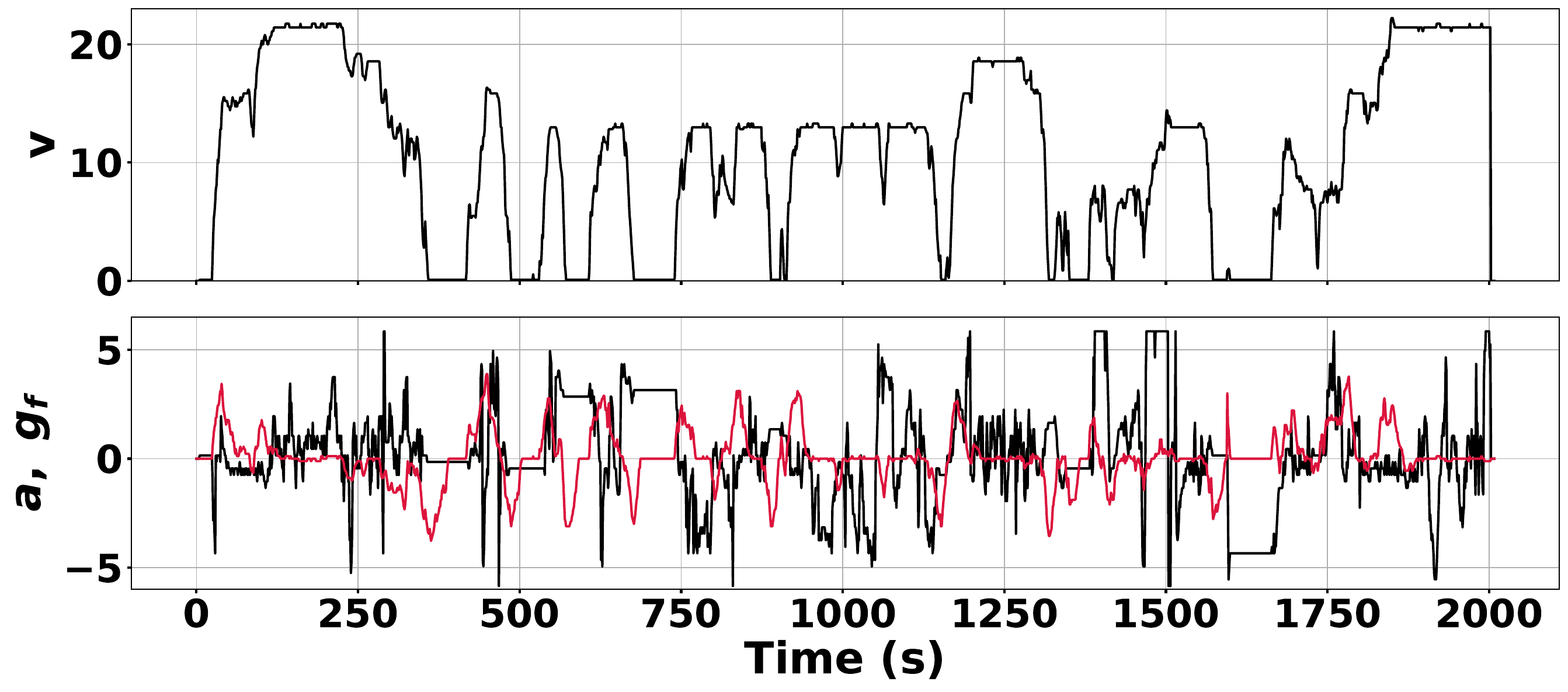}
  \end{subfigure}
  
  \vspace{0.5em}
  \begin{subfigure}[t]{0.45\textwidth}
    \centering
    \caption{MTB DC 1}
    \label{fig:MTB_DC1}
    \includegraphics[width=\textwidth]{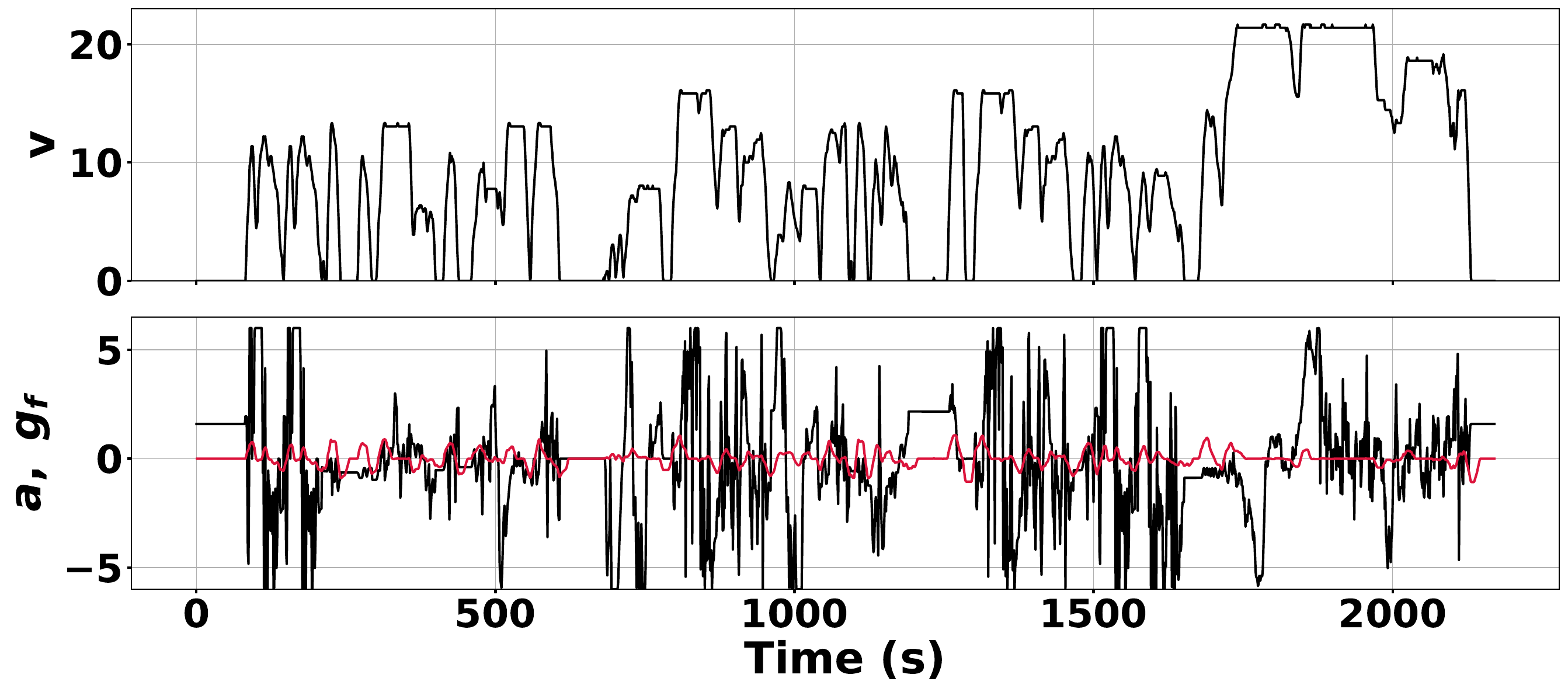}
  \end{subfigure} 
  \hspace{0.5em}
  \begin{subfigure}[t]{0.45\textwidth}
    \centering
    \caption{MTB DC 2}
    \label{fig:MTB_DC2}
    \includegraphics[width=\textwidth]{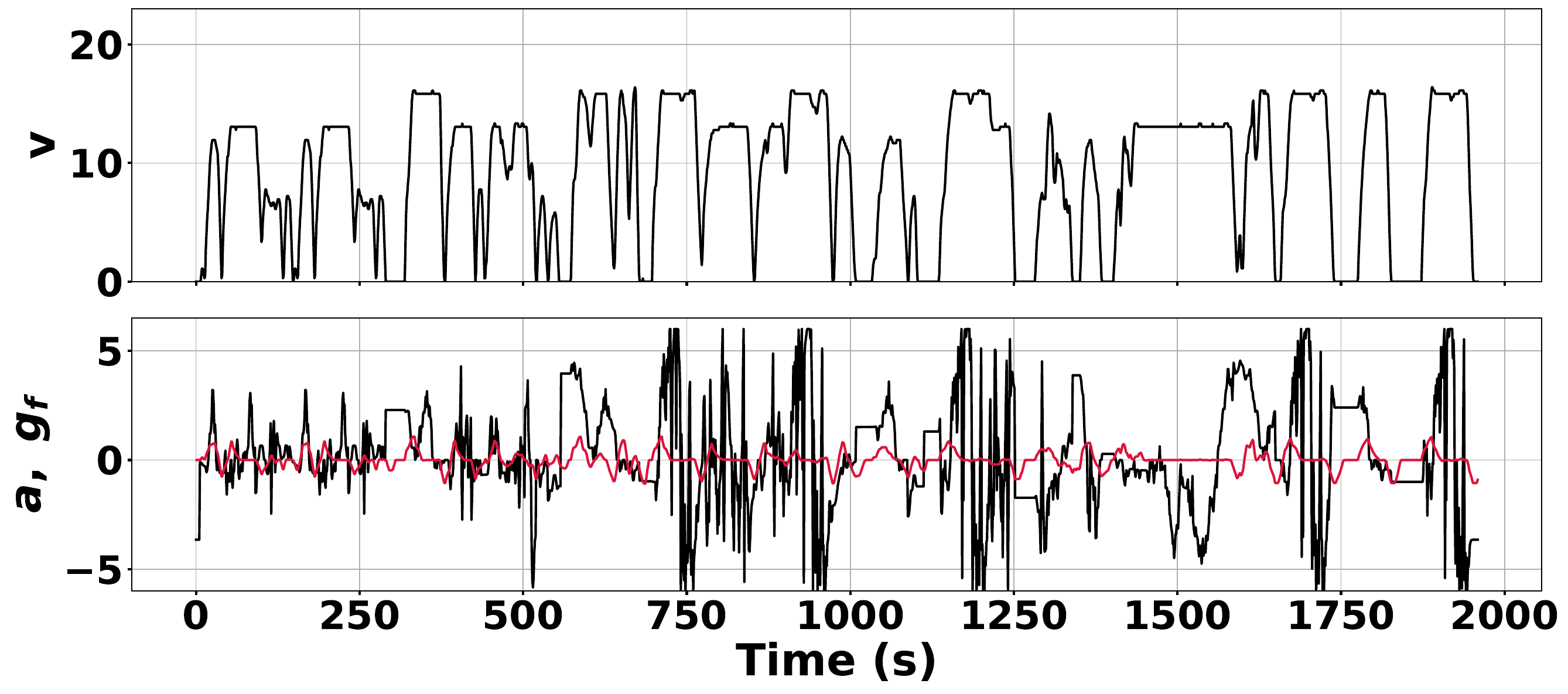}
  \end{subfigure}
  
  \caption{The representative driving cycle using the PIESMC RDCD for different weather conditions. v- speed[m/s], a- acceleration [m/s$^2$], g$_f$ - grade (filtered)[\%]- see Eq. (\ref{eq:savitzky-golay})}
  \label{fig:pirl_driving_cycle}
\end{figure*}

The assessment of the performance of the PIESMC methodology for representative driving cycle construction is described, and a comparison highlighting the advantages and limitations of the PIESMC compared to some of the other methodologies in the literature is provided. The constructed representative driving cycles using the proposed PIESMC method and MCB and MTB methods are also shown in Figure~\ref{fig:pirl_driving_cycle}. To demonstrate the effectiveness of the methods, 50 experiments of complete driving cycles were systematically conducted and analyzed, and the most representative driving cycle constructed by each methodology was depicted.

Figures~\ref{fig:PIESMC_DC1}, ~\ref{fig:PIESMC_DC2} show driving cycle constructed using the PIESMC methodology for DC1 and DC2 (described in section \ref{data_desc}). The proposed approach explicitly incorporates experimental transition dynamics and integrates idling segments by averaging both the duration of idling periods and the intervals between them; consequently, the resulting driving cycle more accurately represents idling behavior while ensuring that acceleration constraints and transitions are considered, leading a realistic deceleration to zero speed and appropriate acceleration to the target velocity while adding the idling sections explicitly.

In contrast, the results obtained from the MCB method (Figures~\ref{fig:MCB_DC1}, \ref{fig:MCB_DC2}) indicate that this method is not effective at capturing the road grade transitions presented in the experimental data which is missing in this methodology because of the high computational and memory costs associated with incorporating detailed road grade information. Furthermore, the MCB method tends to underestimate idling periods due to the inherent randomness of idling occurrences, making it less representative of the actual driving cycles. Finally, as shown in Figure~\ref{fig:MTB_DC1}, \ref{fig:MTB_DC2}, the MTB method is capable of capturing all dimensions of the experimental driving cycle. However, since the MTB approach relies on micro-trip segmentation defined solely by vehicle speed, without considering road grade and acceleration, the resulting constructed driving cycle shows discontinuities in the road grade and acceleration profiles.

To further examine, the frequency content of the road grade profiles \( \alpha(t) \) from the representative driving cycles are examined using Continuous Wavelet Transform (CWT). The CWT is calculated as:
\begin{equation}
W_\alpha(s,\tau) = \frac{1}{\sqrt{s}} \int_{-\infty}^{\infty} \alpha(t) \, \psi\left(\frac{t-\tau}{s}\right)\, dt,
\label{eq:cwt_roadgrade}
\end{equation}
where \( s \) denotes the scale parameter, \( \tau \) is the translation (time-shift) parameter, and \( \psi(t) \) is the mother wavelet, which in this study is chosen to be a Complex Morlet wavelet. The transform \( W_\alpha(s,\tau) \) produces coefficients that depend on both scale (frequency) and translation (time), encapsulating the time-frequency information of the road grade \( \alpha(t) \).

\begin{figure*}[htbp]
  \centering
  \setlength{\abovecaptionskip}{0pt} % Adjust space between caption and figure
  % Second Row
  \begin{subfigure}[t]{0.42\textwidth}
    \centering
    \caption{PIESMC DC1} % (d) on top
    \label{fig:PIRL_Wave_DC1}
    \includegraphics[width=\textwidth]{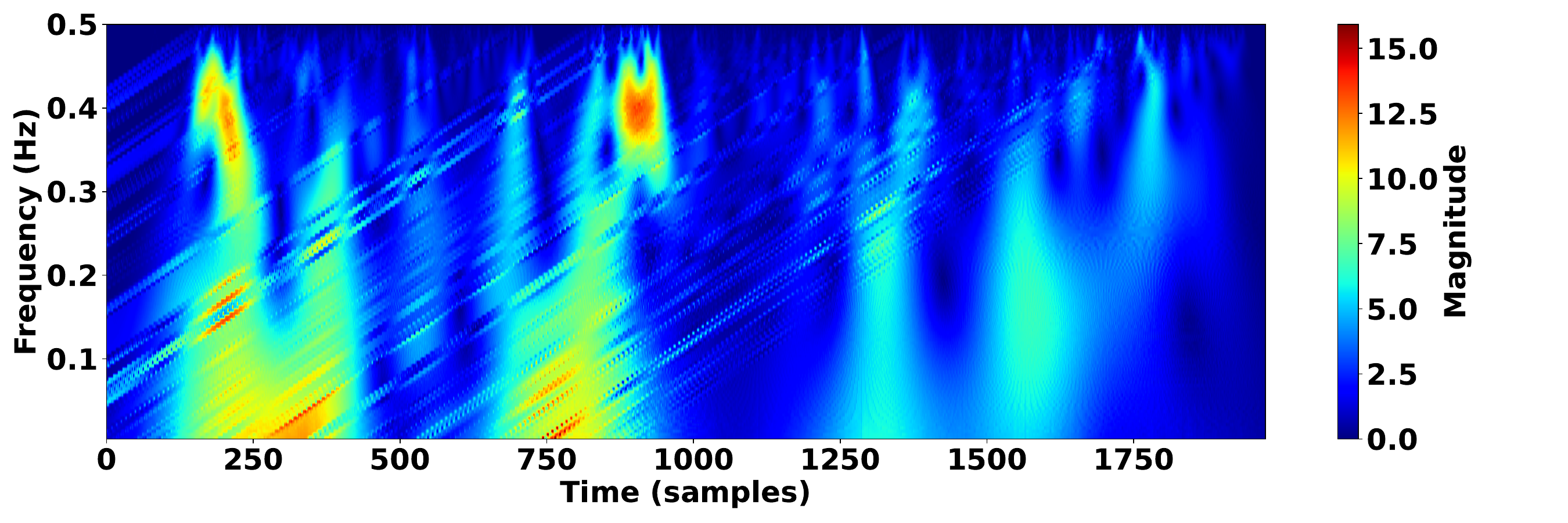}
  \end{subfigure}
  \hspace{0.5em}
  \begin{subfigure}[t]{0.42\textwidth}
    \centering
    \caption{PIESMC DC2} % (e) on top
    \label{fig:PIRL_Wave_DC2}
    \includegraphics[width=\textwidth]{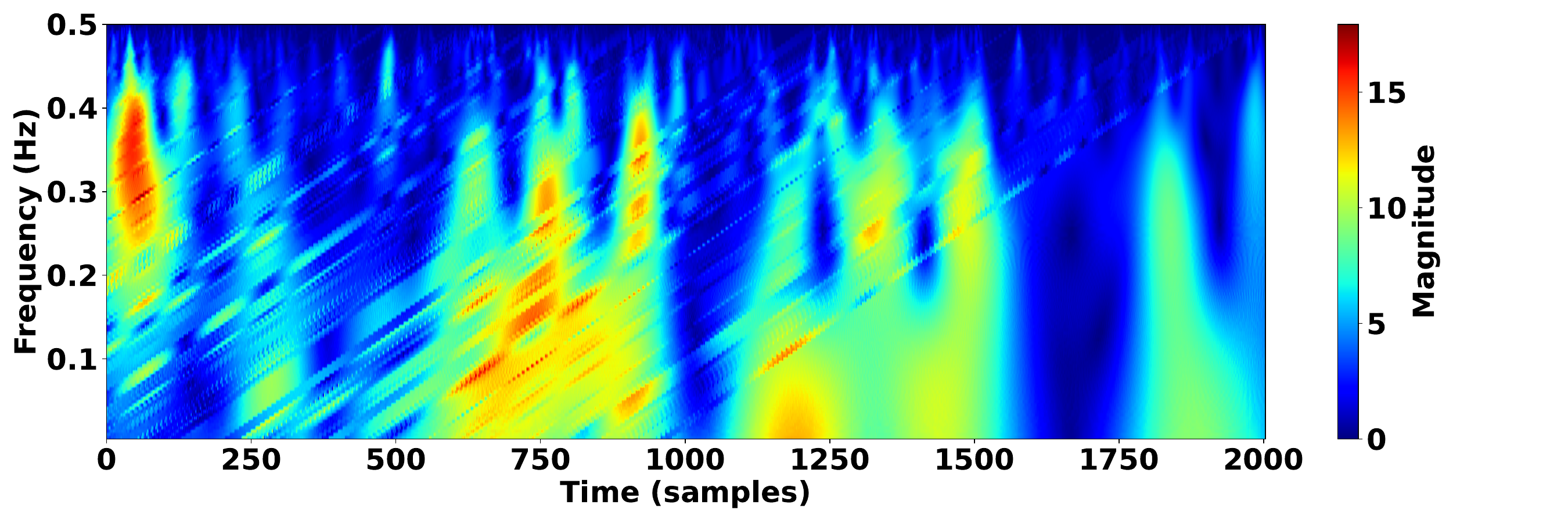}
  \end{subfigure}
  \vspace{0.5em}  
  % \begin{subfigure}[t]{0.45\textwidth}
  %   \centering
  %   \caption{MCB DC1} % (d) on top
  %   \includegraphics[width=\textwidth]{Figures/wavelet_transform_positive_snowy_data_interpolated_grade_CWT_mcb.pdf}
  % \end{subfigure}
  % \hspace{0.5em}
  % \begin{subfigure}[t]{0.45\textwidth}
  %   \centering
  %   \caption{MCB DC2} % (e) on top
  %   \includegraphics[width=\textwidth]{Figures/wavelet_transform_positive_sunny_data_interpolated_grade_CWT_mcb.pdf}
  % \end{subfigure}
  % \vspace{0.5em}  
  \begin{subfigure}[t]{0.42\textwidth}
    \centering
    \caption{MTB DC1} % (d) on top
    \label{fig:MTB_Wave_DC1}
    \includegraphics[width=\textwidth]{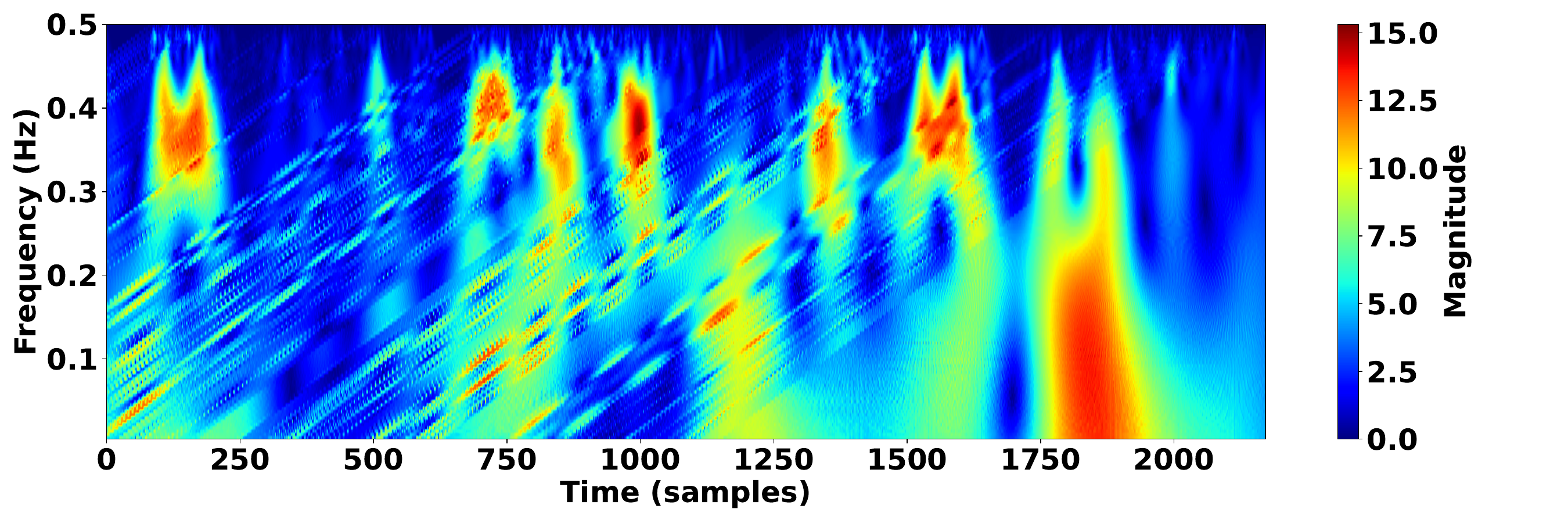}
  \end{subfigure}
  \hspace{0.5em}
  \begin{subfigure}[t]{0.42\textwidth}
    \centering
    \caption{MTB DC2} % (e) on top
    \label{fig:MTB_Wave_DC2}
    \includegraphics[width=\textwidth]{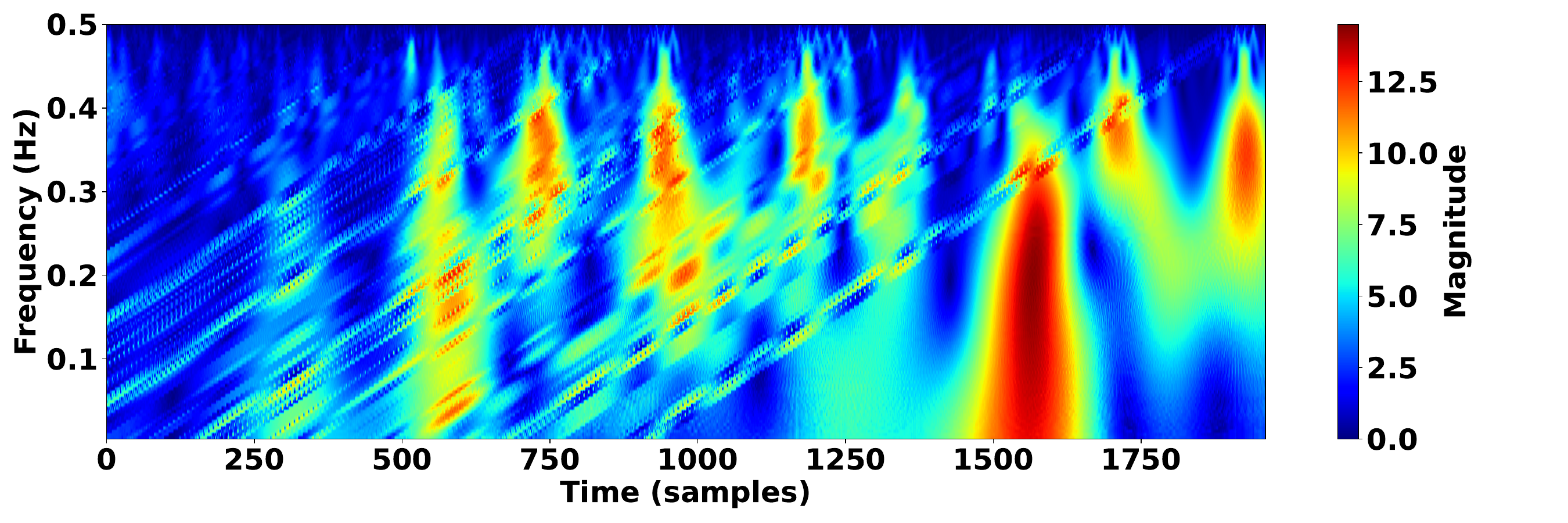}
  \end{subfigure}  
  \caption{Wavelet transformation results of the representative driving cycle for road grade across different methods.}
  \label{fig:wavelet_grade}
\end{figure*}

\noindent The PIESMC method integrates road grade transitions from the experimental data during the construction of the representative driving cycle. As a result, the wavelet transformation of the PIESMC-based grade profiles presented in Figures~\ref{fig:PIRL_Wave_DC1},~\ref{fig:PIRL_Wave_DC2} show a majority of low-frequency components that are consistent with the actual, relatively flat topography of the route. The PIESMC method improves the representativeness of the constructed driving cycles and enhances the reliability of subsequent vehicle modeling and performance assessments. Overall, these findings underscore the critical importance of accurately capturing continuous road grade variations.

In contrast, Figures~\ref{fig:MTB_Wave_DC1},~\ref{fig:MTB_Wave_DC2} present the wavelet transformation results for the MTB method across two distinct experimental datasets. These results showed high-frequency components within the road-grade profiles. This high-frequency content is mainly attributable to the discontinuous nature of the grade profiles generated by the MTB method. Specifically, because the MTB approach constructs the representative driving cycle only based on micro-trip segmentation defined by vehicle speed, it does not explicitly incorporate smooth road grade transitions. As a result, discontinuities in the road grade data result in significant high-frequency artifacts in the wavelet domain. Given that the experimental route is collected within the city of Edmonton, widely recognized for its predominantly flat terrain, the road grade profiles are expected to be relatively smooth and dominated by low-frequency variations. The presence of substantial high-frequency fluctuations in the MTB's results is therefore both unexpected and physically implausible. Such unrealistic oscillations could lead to biased estimates of vehicle emissions and energy consumption if used in further vehicle performance analyses.

\begin{figure*}[htbp]
  \centering
  \setlength{\abovecaptionskip}{0pt} % Adjust space between caption and the figure
  
  \begin{subfigure}[t]{0.45\textwidth}
    \centering
    \caption{{PIESMC DC 1}}
    \label{fig:PI_dist_DC1}
    \includegraphics[width=\textwidth]{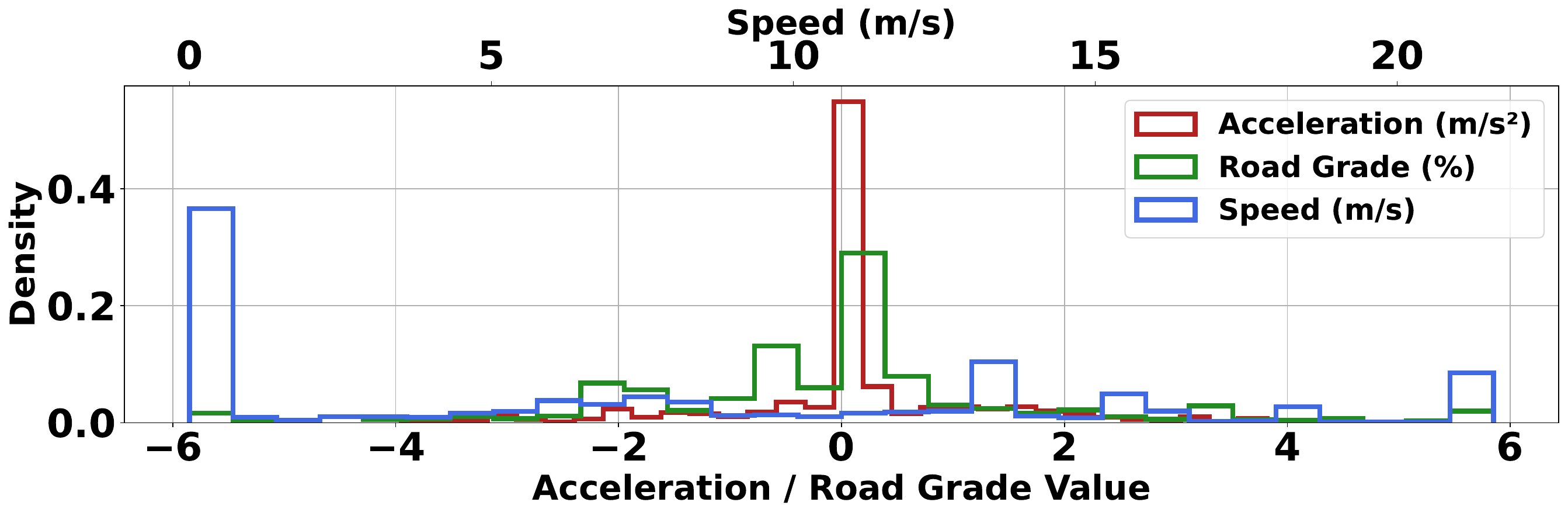}
  \end{subfigure} 
  \hspace{0.5em}
  \begin{subfigure}[t]{0.45\textwidth}
    \centering
    \caption{{PIESMC DC 2}}
    \label{fig:PI_dist_DC2}
    \includegraphics[width=\textwidth]{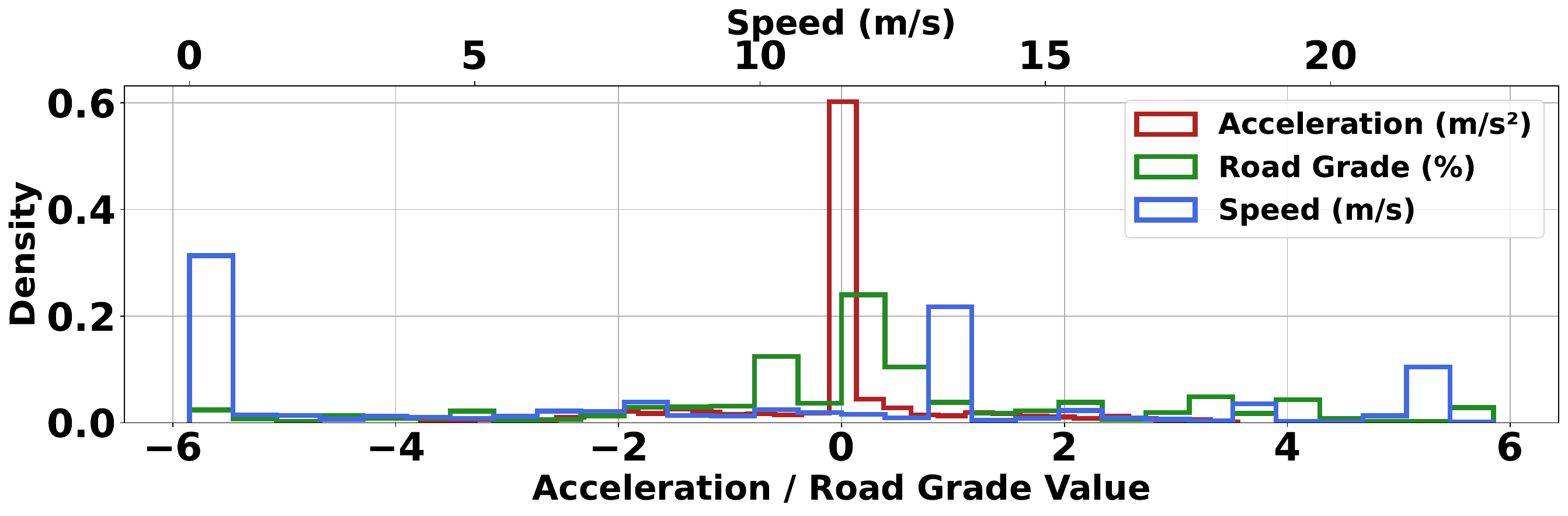}
  \end{subfigure}
    % Second Row
  \vspace{0.5em}
  \begin{subfigure}[t]{0.45\textwidth}
    \centering
    \caption{{MCB DC 1}}
    \label{fig:MCB_dist_DC1}
    \includegraphics[width=\textwidth]{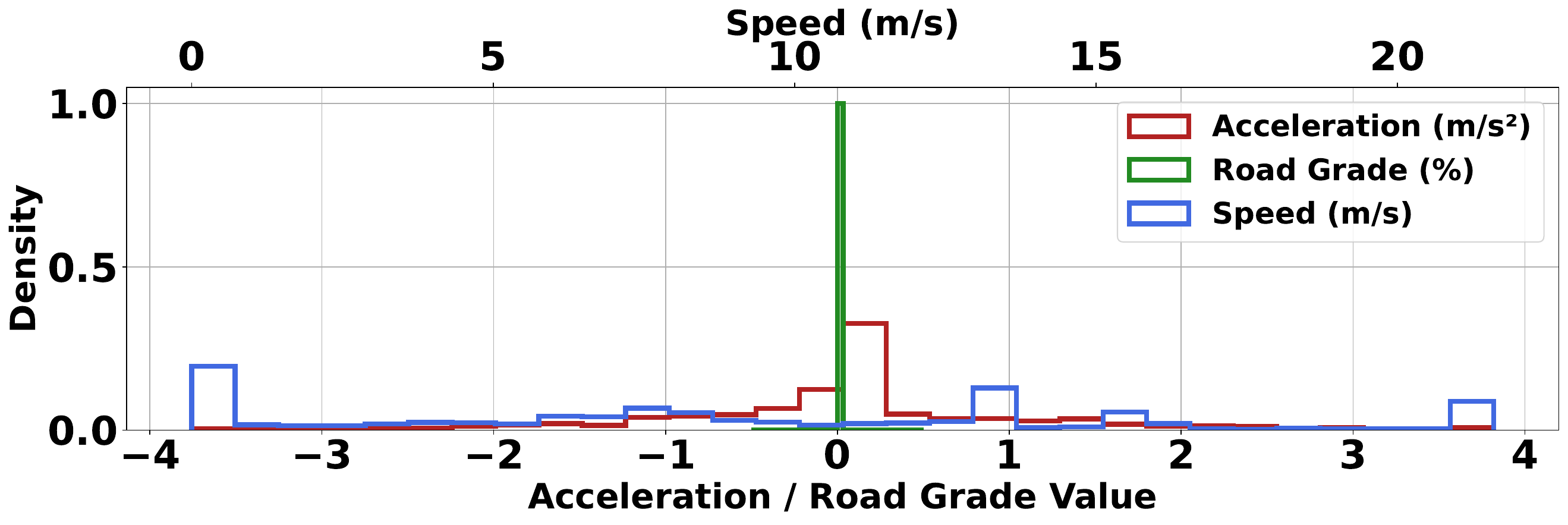}
  \end{subfigure} 
  \hspace{0.5em}
  \begin{subfigure}[t]{0.45\textwidth}
    \centering
    \caption{{MCB DC 2}}
    \label{fig:MCB_dist_DC2}
    \includegraphics[width=\textwidth]{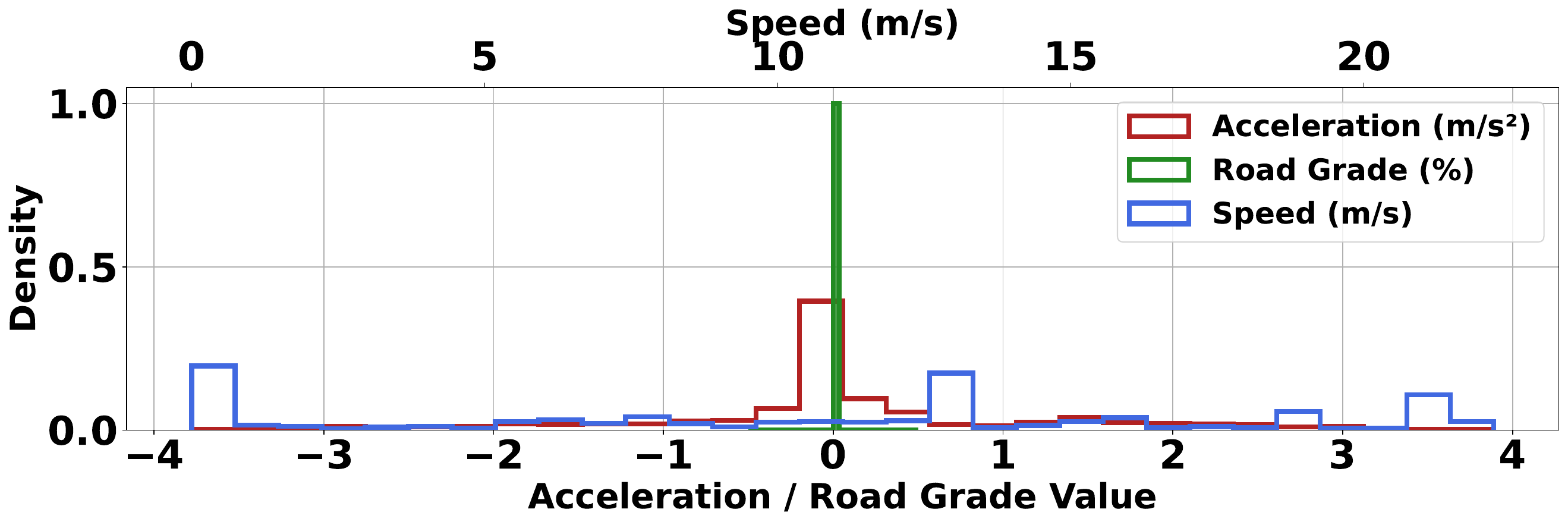}
  \end{subfigure}
    % Second Row
  \vspace{0.5em}
  \begin{subfigure}[t]{0.45\textwidth}
    \centering
    \caption{{MTB DC 1}}
    \label{fig:MTB_dist_DC1}
    \includegraphics[width=\textwidth]{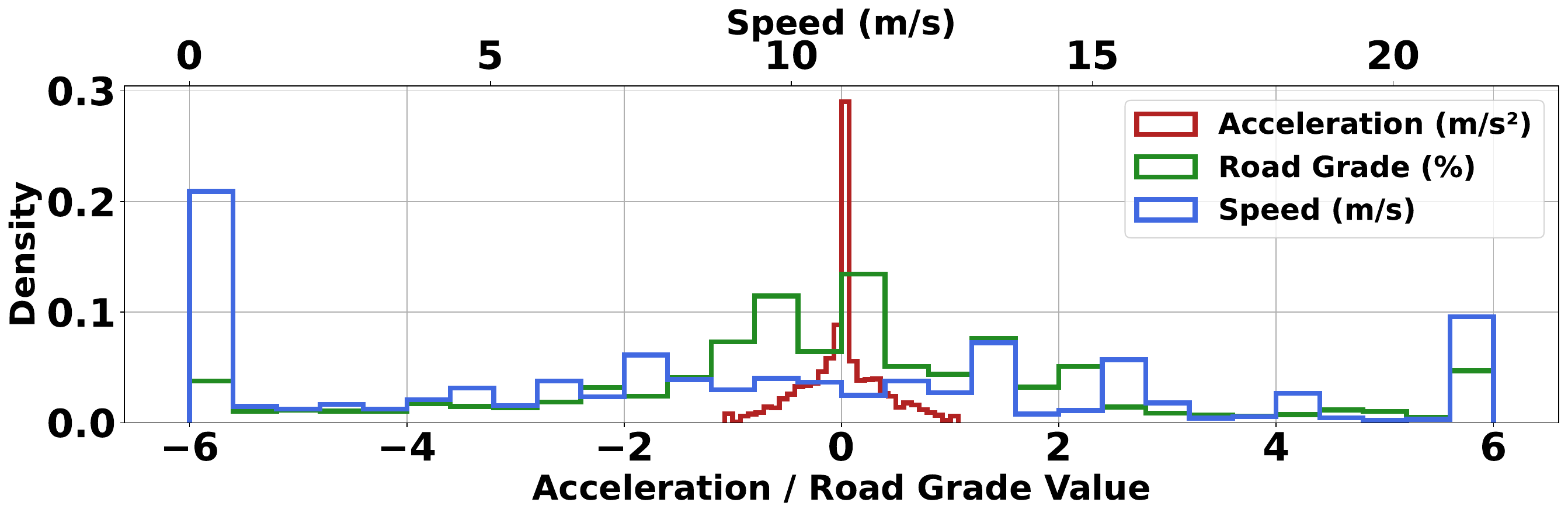}
  \end{subfigure} 
  \hspace{0.5em}
  \begin{subfigure}[t]{0.45\textwidth}
    \centering
    \caption{{MTB DC 2}}
    \label{fig:MTB_dist_DC2}
    \includegraphics[width=\textwidth]{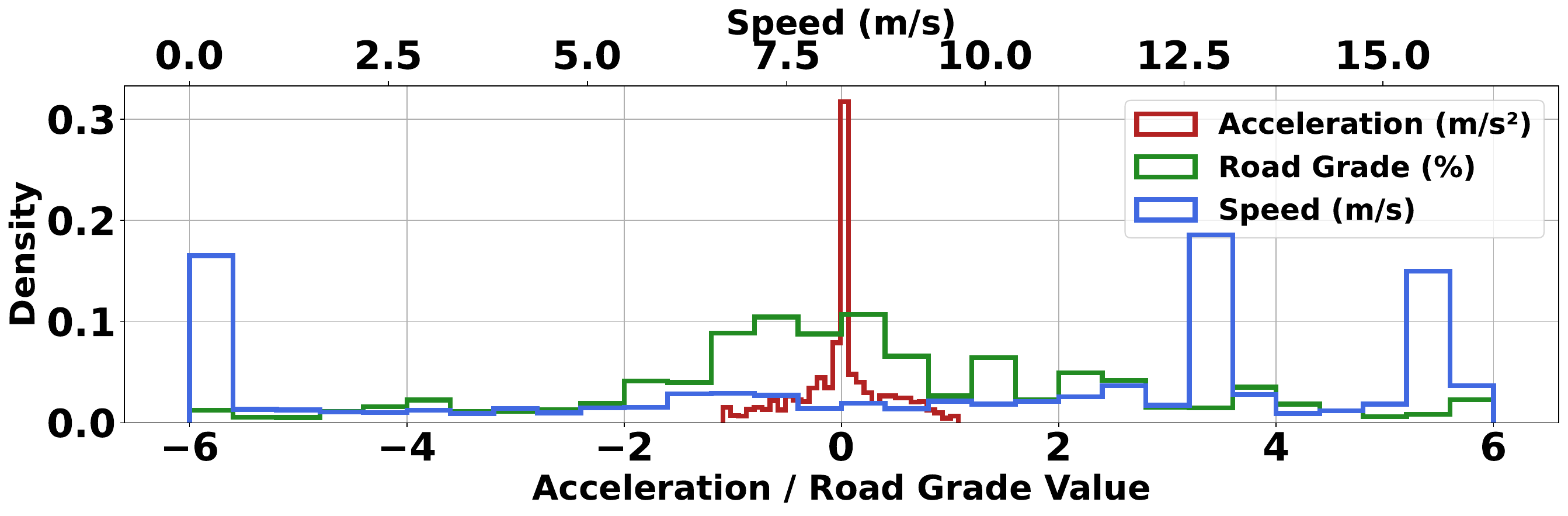}
  \end{subfigure}
  
  \caption{Distribution Statistics for Speed, Acceleration, and Grade by Metric, Method, and Dataset.}
  \label{fig:dist_driving_cycle}
\end{figure*}

The PIESEMC methodology effectively integrates the full range of driving dynamics, including road grade transitions and transient acceleration behavior, to produce representative driving cycles that are both realistic and comprehensive, as shown in Figures~\ref{fig:PI_dist_DC1}, \ref{fig:PI_dist_DC2}. The distributions of vehicle speed, acceleration, and road grade for the representative driving cycles constructed via different methods are shown in Figure~\ref{fig:dist_driving_cycle}. It is evident in Figures~\ref{fig:MCB_dist_DC1}, \ref{fig:MCB_dist_DC2} that the road grade distribution for the MCB method is significantly different from that of the other methods, showing a peak at zero. This peak reflects that the road grade remains essentially zero throughout the constructed driving cycle using the MCB method. Moreover, Figures~\ref{fig:MTB_dist_DC1}, \ref{fig:MTB_dist_DC2} show that the acceleration distribution of the MTB-based driving cycles has a more narrow distribution compared to the other methods. This suggests that the MTB approach underestimates vehicle acceleration, thereby compromising its ability to accurately capture the transient dynamics present in the experimental data. 

\begin{table*}[htbp]
  \centering
  \small
  \caption{Distribution Statistics for Speed, Acceleration, and Grade by Metric, Method, and Dataset.}
  \label{tab:driving_cycle_dist}
  \small
  \setlength{\tabcolsep}{5pt} % Adjust spacing as needed
  \resizebox{0.9\textwidth}{!}{%
    \begin{tabularx}{\textwidth}{ll *{8}{c}}
      \hline
      \textbf{Variable} & \textbf{Metric} & \multicolumn{4}{c}{\textbf{DC1}} & \multicolumn{4}{c}{\textbf{DC2}} \\
      \cline{3-10}
                        &               & \textbf{Actual} & \textbf{MTB} & \textbf{MCB} & \textbf{PIESMC} & \textbf{Actual} & \textbf{MTB} & \textbf{MCB} & \textbf{PIESMC} \\
      \hline
      \multirow{4}{*}{Speed (m/s)} 
        & Min  & 0.00  & 0.00  & 0.00  & 0.00  & 0.00  & 0.00  & 0.00  & 0.00 \\
        & Max  & 21.7 & 21.7 & 21.6 & 21.6 & 23.9 & 16.4 & 22.2 & 22.9 \\
        & Mean & 9.42  & 9.36  & 10.6  & 9.39  & 9.78  & 9.19  & 9.64 & 9.42 \\
        & Std  & 7.00  & 6.83  & 6.60  & 7.28  & 6.85  & 5.70  & 7.23  & 7.55 \\
      \hline
      \multirow{4}{*}{Acceleration (m/s$^2$)}
        & Min  & -2.03 & -1.07 & -3.76 & -3.95 & -2.63 & -1.09 & -3.78 & -3.78 \\
        & Max  & 1.93  & 1.07  & 3.82  & 3.82  & 4.50  & 1.07  & 3.89  & 3.56 \\
        & Mean & 0.00  & 0.00  & 0.04  & 0.09  & 0.02  & 0.00  & 0.06  & -0.06 \\
        & Std  & 0.77  & 0.35  & 1.10  & 1.10  & 0.78  & 0.40  & 1.18  & 1.02 \\
      \hline
      \multirow{4}{*}{Grade (\%)} 
        & Min  & -6.00 & -6.00 & 0.00 & -5.85 & -6.00 & -6.00 & 0.00 & -5.85 \\
        & Max  & 6.00  & 6.00  & 0.00  & 5.85  & 6.00  & 6.00  & 0.00  & 5.85 \\
        & Mean & -0.12 & -0.03 & 0.00 & -0.18 & 0.01  & 0.24  & 0.00  & 0.10 \\
        & Std  & 2.28  & 2.61  & 0.00  & 1.93  & 2.36  & 2.37  & 0.00  & 2.32 \\
      \hline
    \end{tabularx}
  }
\end{table*}

A quantitative comparison by presenting the statistics for these distributions is provided in  Table~\ref{tab:driving_cycle_dist}. The experimental data indicate that the mean speed is 9.42~m/s in DC1 and 9.78~m/s in DC2. In DC1, the MTB method estimates a mean speed of 9.36~m/s, resulting in an absolute error of 0.06~m/s (approximately 0.64\% relative error), whereas, in DC2, it predicts 9.19~m/s, corresponding to an error of 0.59~m/s (around 6.03\% relative error). The MCB method produces a mean speed of 10.6~m/s in DC1, which is an overestimation by 1.18~m/s (about 12.54\% error), but it performs better in DC2 by yielding 9.64~m/s (an error of 0.14~m/s, or roughly 1.43\% relative error). In contrast, the PIESMC method provides mean speeds of 9.39~m/s in DC1 (error of 0.03~m/s, or 0.32\% relative error) and 9.42~m/s in DC2 (error of 0.36~m/s, or 3.68\% relative error), demonstrating its superior ability to capture the central tendency of the speed distribution.

The analysis of acceleration shows that the experimental mean acceleration is nearly 0.00~m/s$^2$ in DC1 and 0.02~m/s$^2$ in DC2. Here, the MTB method returns 0.00~m/s$^2$ for both datasets, and the MCB method estimates 0.04~m/s$^2$ in DC1 and 0.06~m/s$^2$ in DC2. The PIESMC method, however, shows mean accelerations of 0.09~m/s$^2$ in DC1 and -0.06~m/s$^2$ in DC2. Although these differences appear more pronounced given the near-zero experimental values, the absolute deviations of 0.09~m/s$^2$ and 0.08~m/s$^2$ remain acceptable within the context of engineering approximations, especially considering that PIESMC is designed to more accurately capture the natural variability in acceleration.

The most critical comparisons arise in the analysis of road grade profiles where the experimental data show a mean at -0.12\% with standard deviations of 2.28\% in DC1 and 2.36\% in DC2. The MCB method completely fails to capture any variability by yielding a constant 0\% (with zero standard deviation); however, the MTB method underestimates the true spread of road grade fluctuations. In contrast, PIESMC produces mean road grade values of -0.18\% in DC1 and 0.10\% in DC2, along with standard deviations of 1.93\% and 2.32\%, respectively, thereby reflecting the inherent variability more faithfully. This fidelity is especially important for accurately estimating vehicle energy consumption and emissions, as road grade significantly influences vehicle dynamics.

Overall, the PIESMC method exhibits clear advantages over the MTB and MCB methods. It minimizes the error in mean speed (with deviations as low as 0.03~m/s) and maintains acceptable differences in acceleration while accurately reproducing the natural variability in road grade. Such performance is critical for the development of reliable vehicle performance models and environmental impact assessments. The superior fidelity in both central tendencies and dispersion characteristics offered by PIESMC makes it an attractive choice for advanced vehicle dynamics simulations. Future work will focus on further refining these methods and validating their performance across a wider range of driving conditions.

To further assess the performance of the various methods, Table~\ref{tab:kinematic_frag_comp} presents the kinematic fragments obtained from the PIESMC method, experimental measurements, and the MTB and MCB approaches. Regarding the results provided in this table, in terms of the sum error for all kinematic fragments ($\Sigma$ Err.), the proposed PIESMC method outperforms the other methods. In particular, PIESMC achieves error improvements of 35.1\% for DC1 and 57.3\% for DC2 when compared to the MTB method. This significant reduction in error highlights the superior capability of PIESMC in closely replicating the actual kinematic fragments, especially in capturing the transient acceleration and deceleration behaviors, as well as the idling percentage. Moreover, the PIESMC method exhibits minimum deviation in acceleration and deceleration patterns while accurately reflecting the idling percentage. This level of precision was not achieved by either the MTB or MCB methodologies, thereby underscoring the superior effectiveness of the proposed method.

Figure~\ref{fig:dist_driving_cycle} and the results in Table~\ref{tab:kinematic_frag_comp} also indicate that there is a significant gap between the constructed representative driving cycle and the experimental data in terms of acceleration for the MTB method. This shows that the MTB methodology has not performed well in capturing the transient behavior of driving cycles. Furthermore, the results demonstrate that the method underestimates acceleration and deceleration percentages while overestimating cruising and idling percentages. However, regarding the results in Table \ref{tab:kinematic_frag_comp}, it can be concluded that this methodology performs well in terms of average speed without idling since it constructs representative driving cycles based on micro-trips defined by speed.

% The MCB methodology begins with an initial state defined by vehicle speed, acceleration, and road grade. Subsequent states are determined using the SAGSTM, which accounts for experimental data transitions from each state. This approach captures not only speed but also the dynamic transitions in acceleration and road grade, providing a comprehensive representation of driving cycles. To ensure a robust comparison, a maximum of 5,000 driving cycles were generated per experiment or terminated earlier if the SAE in the SAGFD met the predefined threshold, with 50 experiments conducted.

The results for the MCB method in Table~\ref{tab:kinematic_frag_comp} show that this method captures average acceleration and deceleration, in contrast to the MTB methods, which exhibit significant discrepancies in these kinematic fragments. This improved accuracy in representing the transient phases of the driving cycles is attributed to the incorporation of actual transitions in the experimental data for speed and acceleration through the state transition matrix approach. However, Table~\ref{tab:kinematic_frag_comp} highlights that the MCB method has not effectively captured the idling phases of the driving cycles, as there are significant differences between the idling percentage in the experimental data and the constructed driving cycles.

\begin{table*}[htbp]
  \centering
  \small
  \caption{Kinematic Fragments of Constructed RDC for DC1 and DC2.}
  \resizebox{\textwidth}{!}{%
    \begin{tabularx}{\textwidth}{l *{10}{c}}
      \hline
      \textbf{Metric} 
        & \multicolumn{5}{c}{\textbf{DC1}} 
        & \multicolumn{5}{c}{\textbf{DC2}} \\
      \cline{2-11}
                      & \multicolumn{2}{c}{\textbf{Actual}} 
                      & \textbf{MTB} & \textbf{MCB} & \textbf{PIESMC} 
                      & \multicolumn{2}{c}{\textbf{Actual}} 
                      & \textbf{MTB} & \textbf{MCB} & \textbf{PIESMC} \\
      \cline{2-3} \cline{7-8}
                      & \textbf{Mean} & \textbf{Std} 
                      &  &  & 
                      & \textbf{Mean} & \textbf{Std} 
                      &  &  &  \\
      \hline
      $\Bar{a}_p$ [$m/s^2$]   & 1.26   &  0.0353     & 0.483  & 1.34  & 1.21  & 1.26   & 0.101     & 0.471  & 1.28  & 1.29 \\
      $t_{ap}$ [\%]       & 26.9   & 2.19    & 25.3   & 26.8  & 22.9  & 29.3   & 2.07     & 23.1   & 26.5  & 25.4 \\
      $\Bar{a}_n$ [$m/s^2$]  & -1.28  & 0.0268     & -0.477 & -1.32 & -1.25 & -1.26  & 0.0999     & -0.464 & -1.18 & -1.26 \\
      $t_{an}$ [\%]     & 26.6   & 2.01    & 25.2   & 25.3  & 26.0  & 28.4   & 1.95    & 23.2   & 26.9  & 28.2 \\
      $\Bar{V}$ [$m/s$]      & 9.43   & 0.337     & 9.36   & 10.6  & 9.39  & 9.82   & 0.582    & 9.19   & 9.64  & 9.42 \\
      $\Bar{V}_{EI}$ [$m/s$]   & 11.7   & 0.235     & 11.3   & 10.7  & 11.6  & 11.7   & 0.342     & 11.7   & 10.3  & 11.1 \\
      $t_c$ [\%]       & 32.3   & 2.85    & 37.5   & 38.9  & 31.8  & 32.6   & 3.64    & 37.0   & 31.5  & 27.0 \\
      $t_i$ [\%]     & 19.5   & 2.34     & 17.2   & 4.98  & 19.3  & 16.0   & 3.65    & 21.2   & 5.94  & 19.4 \\
      $\Sigma$ Err.       & -      & -     & 180    & 135   & 117   & -      & -     & 213    & 98.7  & 90.9 \\
      Err. Imp. (\%)  & -      & -     & -      & 24.6  & \textbf{35.1}  & -      & -     & -      & 53.6  & \textbf{57.3} \\
      Run Time [$s$]   & -      & -     & 62.9   & 1.35e4 & 436   & -      & -     & 82.4   & 8400  & 254 \\
      \hline
    \end{tabularx}
  }
  \label{tab:kinematic_frag_comp}
\end{table*}

While the MTB method exhibits the most favorable run time (62.9s for DC1 and 82.4s for DC2), its inability to capture the transient behavior leads to several metrics being classified at a higher error level regarding the results in Table~\ref{tab:kinematic_frag_comp}. This shortcoming makes MTB less reliable for analyses where the accurate representation of transient dynamics is critical. In addition, the MCB method is more accurate than MTB in certain aspects; however, it suffers from high computation times (1.35$\times$10$^4$s for DC1 and 8400s for DC2).

Accuracy levels are categorized based on deviation from the mean as follows: Level 1 (within $\pm$1 standard deviation), Level 2 ($\pm$1–2 standard deviations), Level 3 ($\pm$2–3 standard deviations), and Level 4 (greater than $\pm$3 standard deviations). Therefore, it can be seen that the PIESMC method consistently achieves Level 1 or Level 2 accuracy across most metrics, demonstrating that its deviations from the actual data are within one to two standard deviations. In contrast, MTB often registers Level 4 errors for key metrics (such as $\Bar{a}_p$ and $\Bar{a}_n$), and MCB also records several high-level errors, particularly in parameters like $\Bar{V}_{EI}$ and $t_i$. 

In conclusion, despite MTB's efficiency in terms of computational run time, its failure to reliably capture transient dynamics can lead to misleading analyses. On the other hand, PIESMC not only offers superior accuracy but also significantly improves computational efficiency relative to MCB, with run times that are nearly an order of magnitude lower. This combination of accuracy and efficiency establishes the superiority of the PIESMC approach in the construction of representative driving cycles.

\begin{figure*}[htbp]
  \centering
  \setlength{\abovecaptionskip}{0pt} % Adjust space between caption and the figure
  
  \begin{subfigure}[t]{0.45\textwidth}
    \centering
    \caption{{PIESMC DC 1}}
    \includegraphics[width=\textwidth]{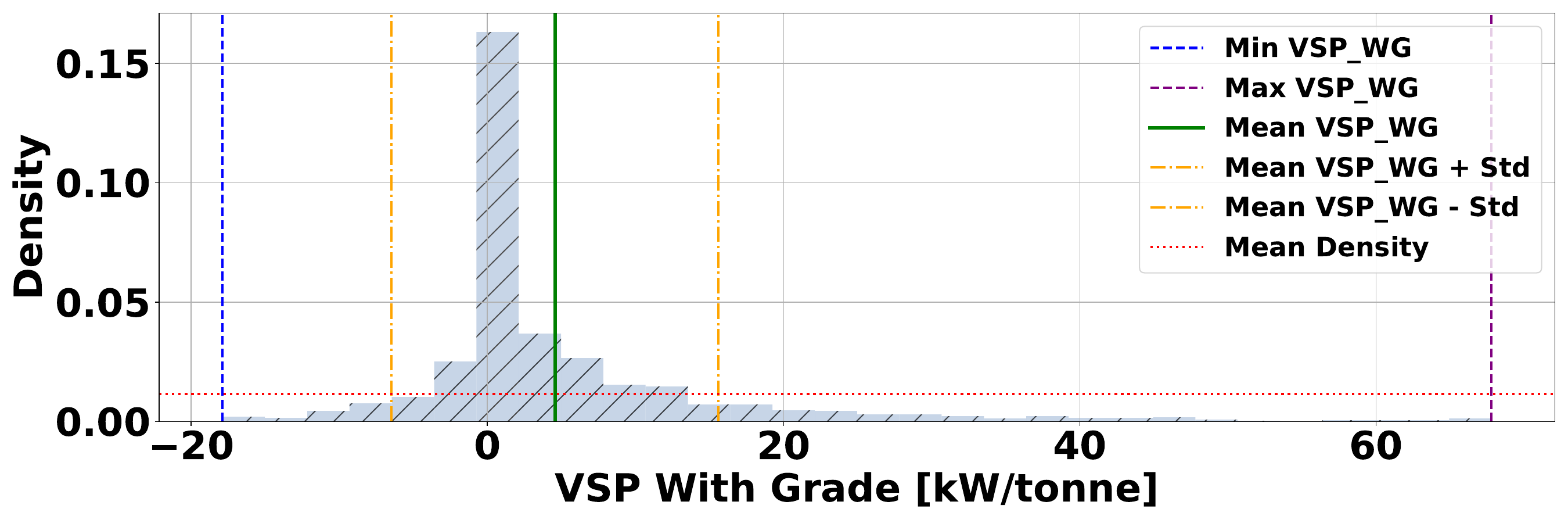}
  \end{subfigure} 
  \hspace{0.5em}
  \begin{subfigure}[t]{0.45\textwidth}
    \centering
    \caption{{PIESMC DC 2}}
    \includegraphics[width=\textwidth]{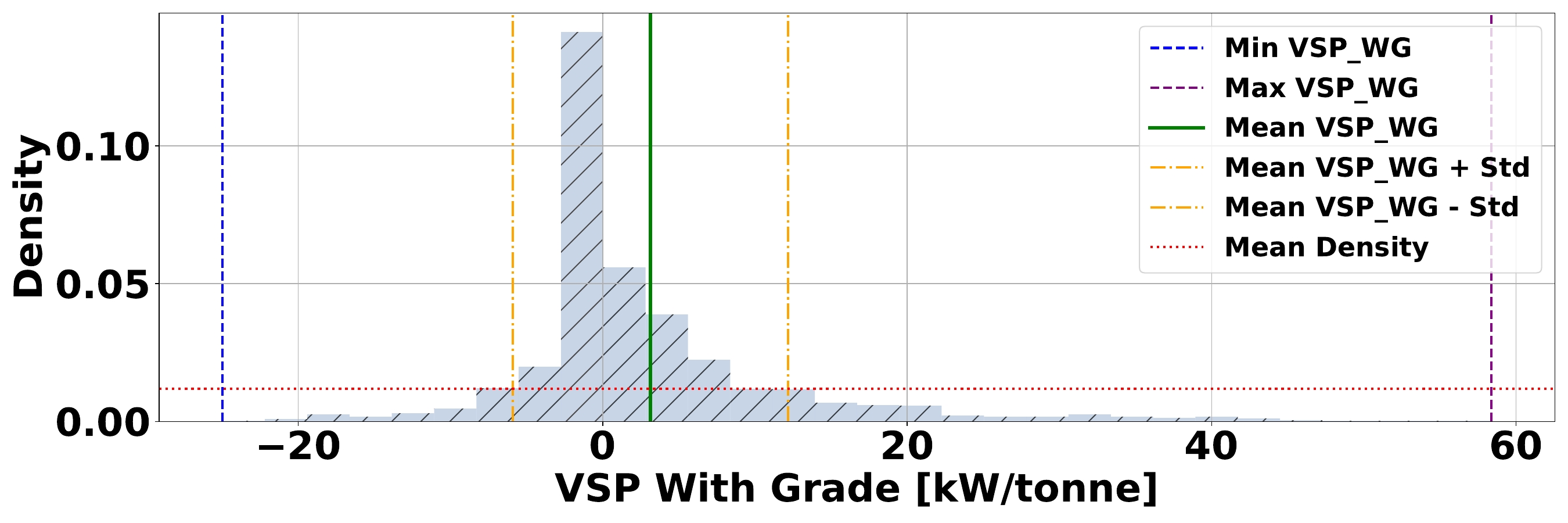}
  \end{subfigure}
    % Second Row
  \vspace{0.5em}
  \begin{subfigure}[t]{0.45\textwidth}
    \centering
    \caption{{MCB DC 1}}
    \includegraphics[width=\textwidth]{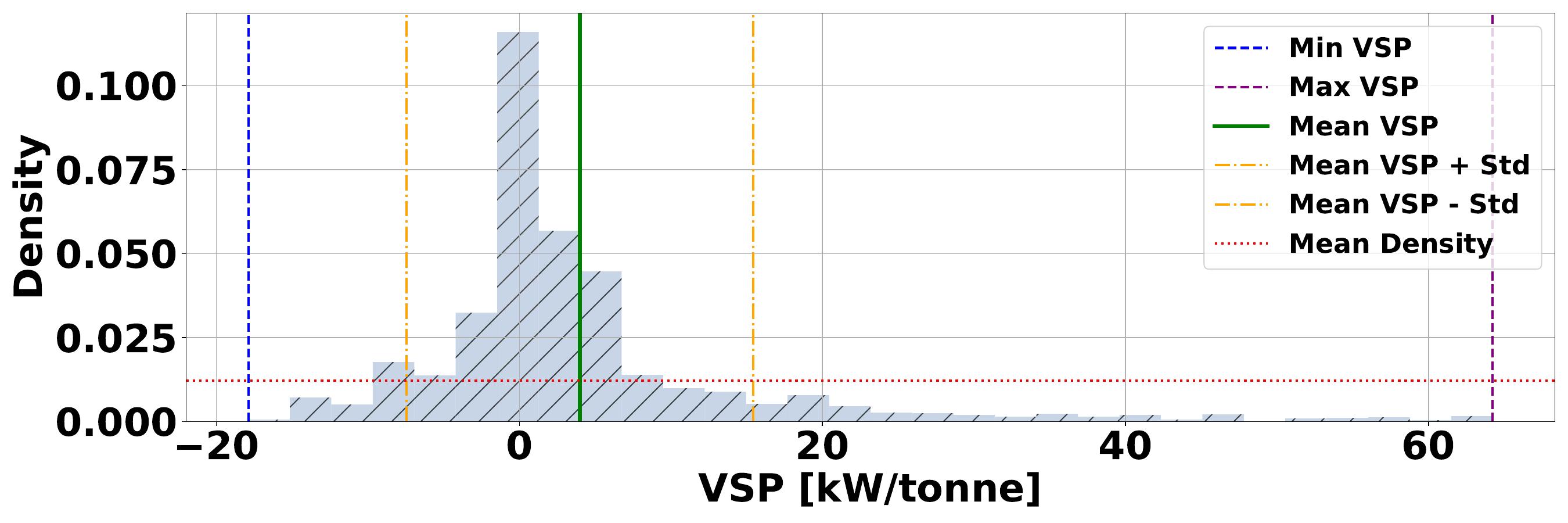}
  \end{subfigure} 
  \hspace{0.5em}
  \begin{subfigure}[t]{0.45\textwidth}
    \centering
    \caption{{MCB DC 2}}
    \includegraphics[width=\textwidth]{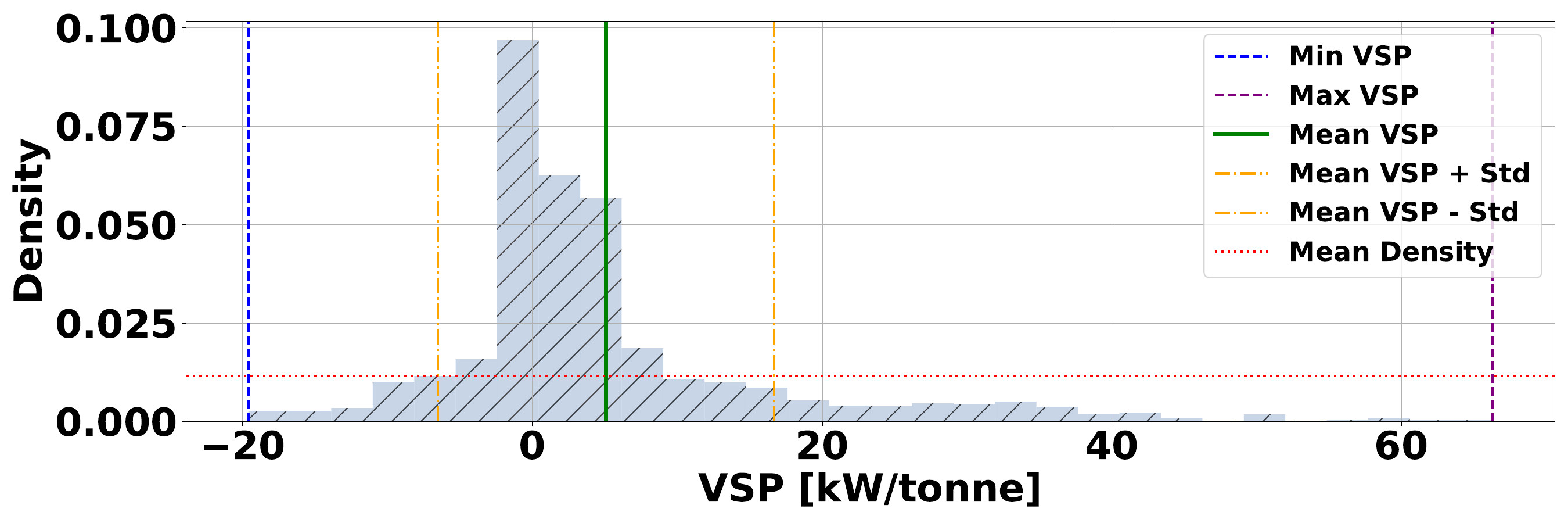}
  \end{subfigure}

    % Second Row
  \vspace{0.5em}
  \begin{subfigure}[t]{0.45\textwidth}
    \centering
    \caption{{MTB DC 1}}
    \includegraphics[width=\textwidth]{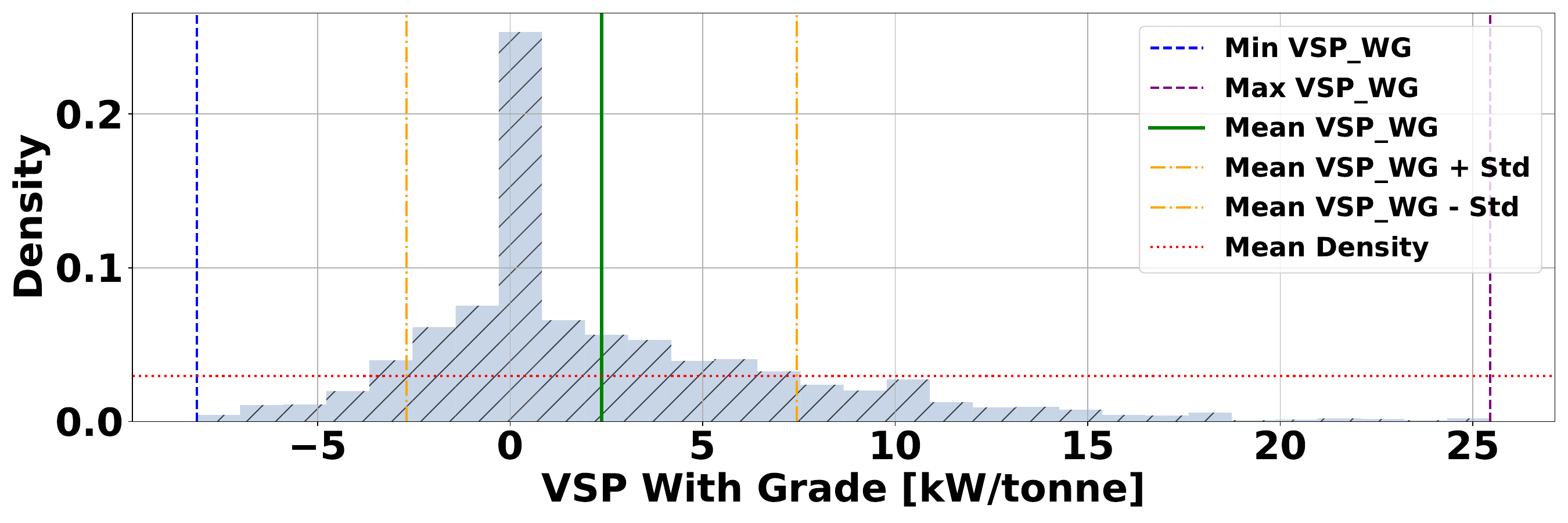}
  \end{subfigure} 
  \hspace{0.5em}
  \begin{subfigure}[t]{0.45\textwidth}
    \centering
    \caption{{MTB DC 2}}
    \includegraphics[width=\textwidth]{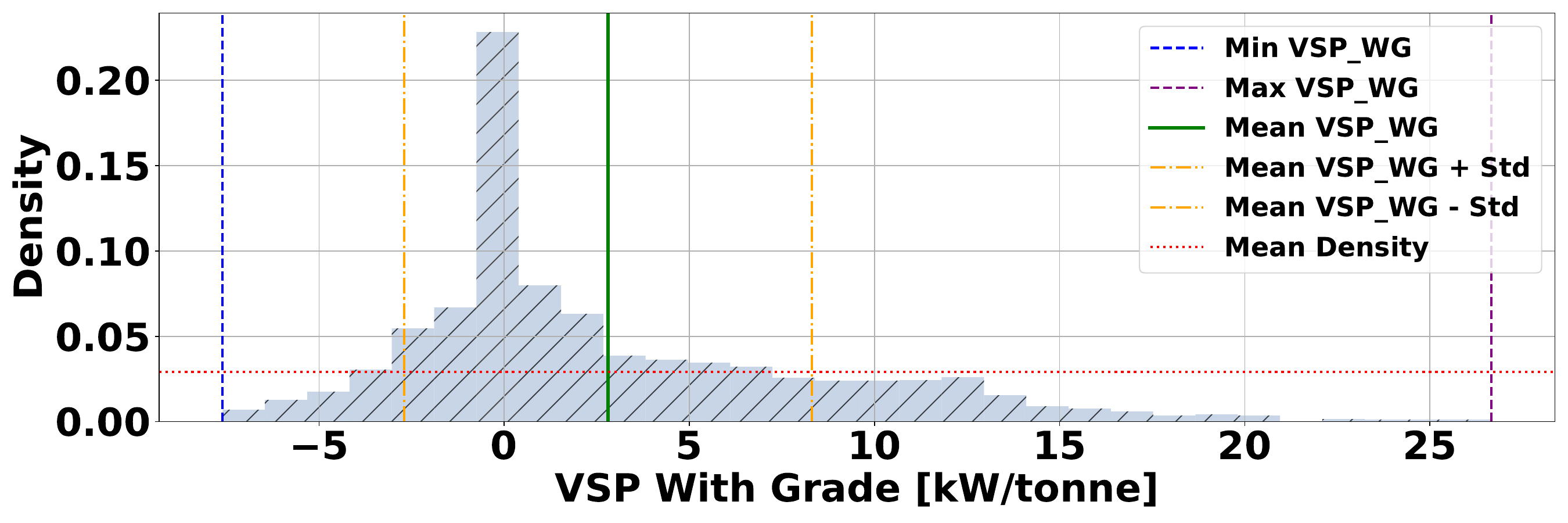}
  \end{subfigure}
  
  \caption{Distribution of VSP across different methods.}
  \label{fig:vsp_distribution}
\end{figure*}

Aside from the kinematic fragments analysis, the VSP distributions for both driving cycles (DC1 and DC2) were computed using the experimental data as well as the MTB, MCB, and PIESMC methods. The VSP distributions for all methods are shown in Figure~\ref{fig:vsp_distribution} and Table~\ref{tab:metrics_methods_datasets} summarizes the key statistical parameters of these VSP distributions, including the minimum, maximum, mean, and standard deviation.

\begin{table*}[htbp]
  \centering
  \small
  \caption{Distribution VSP across different methods.}
  \label{tab:metrics_methods_datasets}
  \small
  \setlength{\tabcolsep}{5pt} % Adjust spacing as needed
  \resizebox{0.95\textwidth}{!}{%
    \begin{tabularx}{0.95\textwidth}{ll *{8}{c}}
      \hline
      \textbf{Variable} & \textbf{Metric} & \multicolumn{4}{c}{\textbf{DC1}} & \multicolumn{4}{c}{\textbf{DC2}} \\
      \cline{3-10}
                        &               & \textbf{Actual} & \textbf{MTB} & \textbf{MCB} & \textbf{PIESMC} & \textbf{Actual} & \textbf{MTB} & \textbf{MCB} & \textbf{PIESMC} \\
      \hline
      \multirow{4}{*}{VSP (kW/ton)} 
        & Min  & -17.19  & -8.13   & -17.89  & -17.87  & -18.23  & -7.60   & -19.61  & -19.96 \\
        & Max  & 41.97   & 25.4  & 64.23   & 67.79   & 60.11   & 26.6  & 66.29   & 58.38 \\
        & Mean & 4.00    & 2.42    & 3.77    & 4.15    & 4.20    & 2.85    & 5.07    & 3.14 \\
        & Std  & 8.41    & 5.48    & 11.04   & 11.04   & 8.16    & 5.93    & 11.60   & 9.04 \\
      \hline
    \end{tabularx}
  }
\end{table*}

For DC1, the PIESMC method achieves a minimum value of \(-17.87\)~kW/ton, which is very close to the experimental minimum of \(-17.19\)~kW/ton. The maximum VSP value obtained with PIESMC is \(67.79\)~kW/ton, exceeding the experimental maximum of \(41.97\)~kW/ton. However, the mean VSP value for PIESMC (4.15~kW/ton) closely approximates the actual mean (4.00~kW/ton), indicating that the central tendency is well captured. For DC2, although the PIESMC method yields a slightly lower mean value (3.14~kW/ton) compared to the experimental mean (4.20~kW/ton), the overall dispersion, as indicated by the standard deviation (9.04~kW/ton for PIESMC versus 8.16~kW/ton for the actual data), remains comparable, effectively replicating the dynamic range observed in the experimental measurements.

The advantages of the PIESMC method are evident when compared to the MTB and MCB approaches. The close match in mean VSP values between PIESMC and the experimental data underscores its accuracy in capturing the central performance trends. Additionally, the comparable standard deviation values suggest that PIESMC reliably represents the VSP distribution, which is critical for robust vehicle performance analysis. The MTB method shows a narrower VSP range, which leads to the underestimation of the vehicle energy consumption and emissions. Conversely, although MCB can sometimes approximate the range more closely, its significantly higher computational expense renders it less practical. In summary, the PIESMC method not only offers enhanced accuracy in reproducing the VSP distribution but also delivers a balanced trade-off between computational efficiency and fidelity, making it a superior approach for driving cycle construction in vehicle design.

\section{Conclusion}\label{conlusion}
A new methodology for constructing representative driving cycles using the PIESMC approach is presented. The primary objective was to develop a method that not only accurately captures the transient dynamics, including acceleration, deceleration, and idling behavior, but integrates critical road grade transitions inherent in experimental driving data. The proposed PIESMC methodology was evaluated through a comprehensive set of experiments, with performance assessed using kinematic and energy-related metrics. The results demonstrated that PIESMC most closely replicates the experimental driving cycle characteristics compared with the other two popular methods. In particular, lower cumulative errors of up to 57.3\% improvement over the MTB method, and maintained a high fidelity in transient acceleration/deceleration profiles and idling percentages. Furthermore, although the MTB method was computationally efficient, its inability to capture high-frequency transient behaviors and smooth road grade transitions limits its utility. Conversely, the MCB method, despite its potential for higher accuracy in certain aspects, incurs high computational costs. A detailed analysis of vehicle-specific power (VSP) distributions further underscored the effectiveness of PIESMC. The method accurately reproduced both the central tendencies and dispersion characteristics of the experimental data, thereby ensuring a robust basis for subsequent vehicle performance assessments and emissions analyses. Additionally, wavelet transformation techniques confirmed that the time-frequency content of the PIESMC-generated cycles aligns well with real-world driving conditions, which was in contrast with the MTB and MCB methods. The PIESMC method presents a balanced trade-off between computational efficiency and representational fidelity, making it a promising tool for advanced vehicle dynamics simulations and environmental impact assessments. Future work will focus on integrating this approach with vehicle dynamics models that could provide deeper insights into energy consumption and emissions under various operating scenarios.

%=========================================================

\section*{Funding}
The authors acknowledge financial support from the Climate Action and Awareness Fund, provided by Environment and Climate Change Canada.  

\section*{Declaration of competing interest}
The authors declare that they have no known competing financial interests or personal relationships that could have appeared to influence the work reported in this paper.

% \section*{Declaration of generative AI and AI-assisted technologies in the writing process}
% During the preparation of this work the author(s) used ChatGPT in order to improve the readability and language of the manuscript. After using this tool/service, the author(s) reviewed and edited the content as needed and take(s) full responsibility for the content of the published article.

% \bibliographystyle{elsarticle-num-names}
% \bibliography{AIforICE_Review_V01,COACH_V01,refHCCInew,ACC19bibliograpgy,ELM_Lit,Misfire_article}

\end{document}